\documentclass{article} 
\usepackage{iclr2025_conference,times}

\usepackage{amsmath,amsfonts,bm}
\usepackage{xspace}

\long\def\red#1{\bgroup\color{red}#1\egroup}

\newcommand{\der} {\mathrm{d}} 

\def\1{\bm{1}}

\newcommand{\rvepsilon}{\blmath{\varepsilon}}
\newcommand{\vtheta}{\blmath{\theta}}
\newcommand{\fvtheta}{\xmath{f_{\vtheta}}}
\newcommand{\Dvtheta}{\xmath{D_{\vtheta}}}

\newcommand{\xmath}[1] {\ensuremath{#1}\xspace}
\newcommand{\blmath}[1] {\xmath{\bm{#1}}} 
\newcommand{\x}{\blmath{x}}
\newcommand{\y}{\blmath{y}}

\def\vs{{\bm{s}}}

\def\vu{{\bm{u}}}

\def\vw{{\bm{w}}}
\def\vx{{\bm{x}}}
\def\vy{{\bm{y}}}
\def\vz{{\bm{z}}}

\def\mA{{\bm{A}}}

\def\mI{{\bm{I}}}

\DeclareMathAlphabet{\mathsfit}{\encodingdefault}{\sfdefault}{m}{sl}
\SetMathAlphabet{\mathsfit}{bold}{\encodingdefault}{\sfdefault}{bx}{n}

\DeclareMathOperator*{\argmin}{arg\,min}

\usepackage{hyperref}
\usepackage{url}
\usepackage{multirow}
\usepackage{adjustbox}
\usepackage[flushleft]{threeparttable}
\usepackage{booktabs}
\usepackage{wrapfig}
\usepackage{algpseudocode}
\usepackage{algorithm}
\usepackage{subcaption}
\usepackage{graphicx}

\title{Patch-Based Diffusion Models Beat Whole-Image Models
for Mismatched Distribution Inverse Problems}

\author{Jason Hu, Bowen Song, Jeffrey A. Fessler, Liyue Shen \\
Department of Electrical and Computer Engineering\\
University of Michigan\\
Ann Arbor, MI 48109, USA \\
\texttt{\{jashu,bowenbw,fessler,liyues\}@umich.edu}
}

\iclrfinalcopy 
\begin{document}

\maketitle

\begin{abstract}
Diffusion models have achieved excellent success in solving inverse problems
due to their ability to learn strong image priors,
but existing approaches require a large training dataset of images
that should come from the same distribution as the test dataset.
When the training and test distributions are mismatched,
artifacts and hallucinations can occur in reconstructed images due to the incorrect priors.
In this work, we systematically study out of distribution (OOD) problems where a known training distribution is first provided.
We first study the setting where only a single measurement obtained from the unknown test distribution is available.
Next we study the setting where a very small sample of data belonging to the test distribution
is available, and our goal is still to reconstruct an image from a measurement that came from the test distribution.
In both settings, we use a patch-based diffusion prior
that learns the image distribution solely from patches.
Furthermore, in the first setting, we include a self-supervised loss
that helps the network output maintain consistency with the measurement.
Extensive experiments show that in both settings,
the patch-based method can obtain high quality image reconstructions 
that can outperform whole-image models
and can compete with methods that have access to large in-distribution training datasets.
Furthermore, we show how whole-image models are prone to memorization and overfitting,
leading to artifacts in the reconstructions, while a patch-based model can resolve these issues.
\end{abstract}

\section{Introduction}

In image processing, inverse problems are of paramount importance
and consist of reconstructing a latent image $\vx$ from a measurement
$\vy = \mathcal{A}(\vx) + \rvepsilon$.
Here, $\mathcal{A}$ represents a forward operator
and $\rvepsilon$ represents random unknown noise.
By Bayes' rule, $\log p(\vx | \vy)$ is proportional to $\log p(\vx) + \log p(\vy | \vx)$,
so obtaining a good prior $p(\vx)$ is crucial for recovering $\vx$
when $\vy$ contains far less information than $\vx$. 
Diffusion models obtain state-of-the-art results for
learning a strong prior and sampling from it,
so similarly competitive results can be obtained when using them to solve inverse problems~%
\citep{MCGchung, chung:23:dps, song2024diffusionblendlearning3dimage, DDNM, kawar2021snips, myown1}.

However, training diffusion models well requires vast amounts of clean training data
\citep{song:21:sbg, ho:20:ddp},
which is infeasible to collect in many applications such as medical imaging
\citep{chung:22:idm, song:22:sip, jalal:21:rcs},
black hole imaging \citep{blackhole_diffusion, feng2024neuralapproximatemirrormaps},
and phase retrieval \citep{myown1, wu:19:orb}. 
In particular, for very challenging inverse problems
such as black hole imaging \citep{blackhole_diffusion}
and Fresnel phase retrieval \citep{fresnel_phase},
no ground truth images are known and one only has a single measurement \y available. 
In other applications such as dynamic CT reconstruction \citep{dynamicCT}
and single photo emission CT \citep{myown2},
obtaining a high quality measurement that can lead
to a reconstruction that closely approximates the ground truth
can be slow or potentially harmful to the patient,
so only a very small dataset of clean images are available.
Thus, in this paper we consider two settings:
the \textit{single measurement} setting in which we are given one measurement \y
whose corresponding \x belongs to a different distribution from the training dataset, 
and the \textit{small dataset} setting in which we are only given a small number of samples \x
that belong to the same distribution as the test dataset.

In recent years, some works have aimed to address these problems
by demonstrating that diffusion models have a stronger generalization ability
than other deep learning methods \citep{jalal:21:rcs},
so slight distribution mismatches between the training data and test data
may not significantly degrade the reconstructed image quality.
However, in cases of particularly compressed or noisy measurements,
as well as when the test data is severely out of distribution (OOD) with a significant domain shift,
an improper choice of training data leads to an incorrect prior
that causes substantial image degradation and hallucinations%
~\citep{blackhole_diffusion, chung_scd}. 
To address these challenges in the single measurement case,
recent works use each measurement \y
to adjust the weights of a diffusion network at reconstruction time
\citet{chung_scd,chung_ddip},
aiming to shift the underlying prior learned by the network
toward the appropriate prior for the latent image in the test case.
However, as the networks have huge numbers of weights,
an intricate and parameter-sensitive refining process of the network is required
during reconstruction to avoid overfitting to the measurement.
Furthermore, there is still a substantial gap in performance between methods
using an OOD prior and methods using an in-distribution prior.
Finally, these methods have only been tested in medical imaging applications \citep{chung_scd, chung_ddip}.
On the other hand, in the small dataset case,
various methods \citep{moon2022finetuning, soda_finetune} have been devised
to fine-tune a diffusion model on an OOD dataset,
but these methods still require several hundred images
and have not used the fine-tuned network to solve inverse problems.

Patch-based diffusion models have shown success both for image generation
\citep{wang2023patch, ding2023patched}
and for inverse problem solving \citep{hu2024learningimagepriorspatchbased}.
In particular, the method of \citet{hu2024learningimagepriorspatchbased} involves training networks
that take in only patches of images at training and reconstruction time,
learning priors of the entire images from only priors of patches.
In cases of limited data,
\citet{hu2024learningimagepriorspatchbased} shows that patch-based diffusion models
outperform whole image models for solving certain inverse problems.
These works motivate our key insight that patch-based diffusion priors
potentially obtain stronger generalizability than whole-image diffusion priors
for both the single measurement setting and the small dataset setting due to a severe lack of data.
Inspired by this, we propose to utilize patch-based diffusion models
to tackle the challenges arising from mismatched distributions and lack of data in a unified way.
We first develop a method to take a network trained on patches of a mismatched distribution
and adjust it on the fly in the single measurement setting.
We also show how in the small dataset setting, fine-tuning a patch-based network
results in a much better prior than fine-tuning a whole-image network,
leading to higher quality reconstructed images.

In summary, our contributions are as follows:
\begin{itemize}
\item
We integrate the patch-based diffusion model framework
with the deep image prior (DIP) framework
to correct for mismatched distributions in the single measurement setting.
Experimentally, we find this approach beats using whole-image models
in terms of quantitative metrics and visual image quality in image reconstruction tasks,
as well as achieving competitive results with methods using in-distribution diffusion models.
\item
We show that in the small dataset setting,
fine-tuning patch-based diffusion models is much more robust than whole-image models
and very little data is required to obtain a reasonable prior for solving inverse problems.
\item
We demonstrate experimentally that when fine-tuning on very small datasets,
whole image diffusion models are prone to overfitting and memorization,
which severely degrades reconstructed images,
while patch-based models are much less sensitive to this problem.
\end{itemize}

\section{Background and Related Work}
\label{section2}

\textbf{Diffusion models and inverse problems.}
In a general framework,
diffusion models involve the forward stochastic differential equation (SDE)
\begin{equation}
\der \vx = -\frac{\beta(t)}{2} \, \vx \, \der t + \sqrt{\beta(t)} \, \der \vw,
\end{equation}
where $t \in [0, T]$, $\vx(t) \in \mathbb{R}^d$, 
and $\beta(t)$ is the noise variance schedule of the process. 
This process adds noise to a clean image and ends with an image indistinguishable from Gaussian noise ~\citep{song:21:sbg}.
Thus, the distribution of $\vx(0)$ is the data distribution
and the distribution of $\vx(T)$ is (approximately) a standard Gaussian.
Then the reverse SDE has the form~\citep{ANDERSON1982313}: 
\begin{equation}
\der \vx = \left( -\frac{\beta(t)}{2} - \beta(t) \nabla_{\vx_t} \log p_t(\vx_t) \right) \, \der t
+ \sqrt{\beta(t)} \, \der \overline{\vw}.
\end{equation}
Score-based diffusion models involve training a neural network to learn the score function
$\nabla_{\vx_t} \log p_t(\vx_t)$,
from which one can start with noise and run the reverse SDE
to obtain samples from the learned data distribution.

When solving inverse problems, it is necessary to instead sample from $p(\vx_T | \vy)$,
so the reverse SDE becomes 
\begin{equation} \label{eq: revSDE}
\der \vx = \left( -\frac{\beta(t)}{2}
- \beta(t) \nabla_{\vx_t} \log p_t(\vx_t | \vy) \right) \, \der t
+ \sqrt{\beta(t)} \, \der \overline{\vw}.
\end{equation}
Unfortunately, the term $\log p_t(\vx_t | \vy)$ is difficult to compute
from the unconditional score  $\nabla_{\vx_t} \log p_t(\vx_t)$ alone.
\citet{i2sb}, \citet{cddb}, and \citet{visionpatch} among others
proposed directly learning this conditional score $\nabla_{\vx_t} \log p_t(\vx_t | \vy)$ instead.
However, this process requires paired data $(\vx, \vy)$
between the image domain and measurement domain for training,
instead of just clean image data.
Furthermore, the learned conditional score function
is suitable only for the particular inverse problem for which it was trained,
limiting its flexibility.

For greater generalizability,
it is desirable to apply the unconditional score $\nabla_{\vx_t} \log p_t(\vx_t)$
to be able to solve a wide variety of inverse problems.
Thus, many works
have been proposed to approximate the conditional score in terms of the unconditional one 
\citep{DDNM, chung:23:dps, chung2024decomposeddiffusionsampleraccelerating, DDRM}.
Notably, \citet{peng2024improvingdiffusionmodelsinverse}
unified various diffusion inverse solvers (DIS) into two categories:
the first consists of direct approximations to $p_t(\vy | \vx_t)$,
and the second consists of first approximating $\mathbb{E}[\vx_0 | \vx_t, \vy]$
(typically through an optimization problem balancing the prior and measurement)
and then applying Tweedie's formula \citep{efron2011tweedies} to obtain 
\begin{equation}
    \nabla \log p_t(\vx_t | \vy) = \frac{\mathbb{E}[\vx_0 | \vx_t, \vy] - \vx_t}{\sigma_t^2},
\end{equation}
where $\sigma_t$ is the noise level of $\vx_t$.
All of these methods require a large quantity of clean training data
that should come from the distribution $p(\vx)$
whose score is to be learned, which may not be available in practice.

\textbf{Methods without clean training data.}
When no in-distribution data is available, one approach is to use traditional methods
that do not require any training data,
such as total variation (TV) \citep{ctlowrank}
or wavelet transform \citep{daubechies:92:tlo} regularizers
that encourage image sparsity.
More recently, plug and play (PnP) methods have risen in popularity
\citep{xiaojian2, Sreehari.etal2016, 9026811, hong2024ppnp};
these methods use a denoiser to solve general inverse problems. 
Although these methods often use a trained denoiser,
\citet{ryu2019plugandplaymethodsprovablyconverge}
found that using an off-the-shelf denoiser such as block matching 3D \citep{dabov2006bm3d}
can yield competitive results.
Nevertheless, with the rise of deep learning in image processing applications,
methods that harness the power of these tools may be desirable.

The deep image prior (DIP) is an extensively studied self-supervised method
that is popular when no training data is available
and reconstruction from a single meausurement \y is desired.
The method consists of training a network \fvtheta using the loss function 
\begin{equation} \label{eq:dip}
    L(\vtheta) = \| \vy - \mathcal{A}(\fvtheta(\vz)) \|_2^2, \quad \vz \sim \mathcal{N}(0, \mI),
\end{equation}
so that $\fvtheta(\vz)$ produces the reconstruction.
Although the neural network acts as an implicit regularizer
whose output tends to lie in the manifold of clean images,
DIP is prone to overfitting \citep{dip_first}.
Various methods have been proposed
involving early stopping, regularization, and network initialization
\citep{xiaojian1, jo2021rethinkingdeepimageprior, DIP_CT}.
Nevertheless, the method is very sensitive to parameter selection and implementation
and can take a long time to train \citep{jo2021rethinkingdeepimageprior}. 

Most DIS methods learn a prior from a large collection of clean in-distribution training images,
but recently \citet{chung_scd} and \citet{chung_ddip}
proposed self-supervised diffusion model methods that are based off the DIP framework.
These methods involve alternating between the usual reverse diffusion update step
to gradually denoise the image
and a network refining step in which the score network parameters are updated via the loss function 
\begin{equation}
    L(\vtheta) = \| \vy - \mathcal{A}( \text{CG}(\hat{\vx}_{0|t}(\vx_t; \vtheta))) \|_2^2
\end{equation}
where conjugate gradient (CG) descent is used to enforce data fidelity. 
This CG step consists of solving an optimization of the form
\begin{equation}
\label{eq:cg-step}
    \argmin_{\vx} \frac{\gamma}{2} \| \vy - \mathcal{A}(\vx) \|_2^2
    + \frac{1}{2} \| \vx - \hat{\vx}_{0|t}\|_2^2,
\end{equation}
where $\gamma$ is a tradeoff parameter controlling the strength of the prior versus the measurement. 
Crucially, these methods introduce an additional LoRA module \citep{lora} to the network
and the original network parameters are frozen when backpropagating the loss,
which helps to avoid overfitting
the whole-image model.
Nevertheless, many technical tricks are required \citep{chung_ddip}
involving noisy initializations and early stopping to obtain good results and avoid artifacts.
Our patch-based model avoids this overfitting issue.

\textbf{Diffusion model fine-tuning.}
In the small dataset setting,
various fine-tuning methods exist
to shift the underlying prior learned by a score network
away from a mismatched distribution and toward a target distribution.
Given a pretrained diffusion network on a mismatched distribution,
\cite{moon2022finetuning}, \cite{soda_finetune}, and \cite{zhu2024_finetune} among others
have studied ways to fine-tune the network to the desired dataset.
These methods generally involve freezing certain layers of the original network,
appending extra modules that contain relatively few weights,
or modifying the loss function to capture details that differ greatly between distributions.
However, these methods usually still require thousands of images from the desired distribution
and focus on image generation. 
When solving inverse problems,
the reconstructed image should be consistent with the measurement \y, reducing the number of degrees of the freedom for the image compared to generation,
so with proper fine-tuning the data requirement should be lower.

\section{Methods}
\label{methods}

\subsection{Patch-based prior}
\begin{wrapfigure}{r}{0.5\textwidth}
\centering
\includegraphics[width=0.48\textwidth]{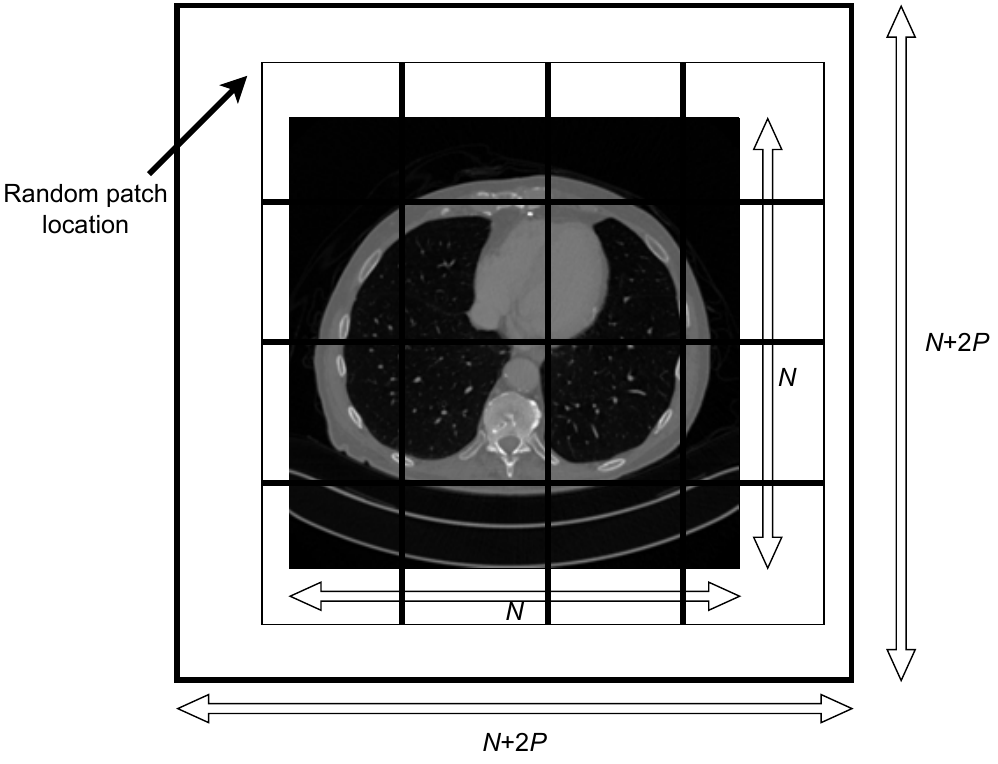}
\caption{Schematic for zero padding and partitioning image into patches.
Each index $i$ represents one of the $M^2$ possible ways to choose a patch location.}
\label{fig:patch_partition}
\end{wrapfigure}
We adapt the patch-based diffusion model framework of \citet{hu2024learningimagepriorspatchbased};
we zero pad the image by an amount $P$ on each side
and analyze the distribution of the resulting image $\vx$. 
Then we assume the true underlying data distribution takes the form 
\begin{equation} \label{eq:padis_distribution}
    p(\vx) = \prod_{i=1}^{M^2} p_{i, B}(\vx_{i,B}) \prod_{r=1}^{(k+1)^2} p_{i,r}(\vx_{i,r})/Z,
\end{equation}
where $\vx_{i,B}$ represents the aforementioned bordering region of $\vx$
that depends on the specific value of $i$,
$p_{i, B}$ is the probability distribution of that region,
$\vx_{i,r}$ is the $r$th $P \times P$ patch
when using the partitioning scheme corresponding to the value of $i$,
$p_{i,r}$ is the probability distribution of that region,
and $Z$ is an appropriate scaling factor.

For training, we use a neural network $\Dvtheta(\vx, \sigma_t)$
that accepts a noisy image $\vx$ and the noise level $\sigma_t$.
For each patch, we define the $x$ positional array
as the 2D array consisting of the $x$ positions of each pixel of the image scaled between -1 and 1.
To allow the network to learn different patch distributions at different locations in the image,
we extract the corresponding patches of these positional arrays
and concatenate them along the channel dimension of the noisy image patch
and treat the entire array as the network input. 
Since we are using a patch-based prior, we perform denoising score matching on patches of an image instead of the whole image. Hence, the training loss is given by 
\begin{equation} \label{DSM}
\argmin_{\vtheta}
\mathbb{E}_{t \sim \mathcal{U}(0,T)} \mathbb{E}_{\vx \sim p(\vx)}
\mathbb{E}_{\rvepsilon \sim \mathcal{N} (0, \sigma_t^2 I)}
\| \Dvtheta(\vx + \rvepsilon, \sigma_t) - \vx \|_2^2,
\end{equation}
where $\vx \sim p(\vx)$ represents a patch drawn from a sample of the training dataset, $\sigma_t$ is a predetermined noise schedule, and $\mathcal{U}$ represents the uniform distribution.

\subsection{Single measurement setting}
Consider the first case where only the measurement \y is given,
and no in-sample training data is available.
For each specific measurement \y,
the DIP framework optimizes
the network parameters \vtheta
via the self-supervised loss
\eqref{eq:dip}
from the predicted reconstructed image.
Diffusion models provide a prediction of the reconstructed image at each timestep:
namely, the expectation of the clean image $\mathbb{E}[\vx_0 | \vx_t]$
is approximated by the denoiser $\Dvtheta(\vx_t)$ via Tweedie's formula.
Then the expectation conditioned on the measurement 
$\mathbb{E}[\vx_0 | \vx_t, \vy]$ can be obtained
through one of many methods of enforcing the data fidelity constraint.

We begin with the unconditional expectation by leveraging the patch-based prior. 
Following \eqref{eq:padis_distribution}, we apply Tweedie's formula
to express the denoiser of $\vx$ in terms of solely the denoisers of the patches of $\vx$.
Because the outermost product is computationally very expensive,
in practice we approximate $\Dvtheta(\vx)$
using only a single randomly selected value of $i$
for each denoiser evaluation:
\begin{equation} \label{eq:denoiser_stochastic}
    \Dvtheta(\vx) \approx D_{i,B}(\vx_{i,B}) + \sum_{r=1}^{(k+1)^2} D_{i,r}(\vx_{i,r}).
\end{equation}
By definition, $D_{i,B}(\vx_{i,B})=0$
and we compute each $D_{i,r}(\vx_{i,r})$ with the network.
Note that \eqref{eq:denoiser_stochastic}
provides an \textit{unconditional} estimate of the clean image; 
to obtain a conditional estimate $\Dvtheta(\vx_t | \vy)$ of the clean image,
we run $M$ iterations of the conjugate gradient descent algorithm
for minimizing
$\| \mA \x - \y \|_2$,
initialized with the unconditional estimate \citep{chung2024decomposeddiffusionsampleraccelerating}.

The image that is being reconstructed
might not come from the distribution of the training images.
Hence, the estimate $\Dvtheta(\vx_t | \vy)$ may be far from the true denoised image.
Thus, we use \y to update the parameters of the network
in a way such that $\Dvtheta(\vx_t | \vy)$ becomes more consistent with the measurement: 
\begin{equation}
\label{eqn-ss}
    \vtheta \leftarrow \argmin_{\vtheta} \| \y - \mA \, \Dvtheta (\x_t | \y)\|_2^2.
\end{equation}

\begin{wrapfigure}{r}{0.5\textwidth}
\begin{minipage}{\linewidth}
\begin{algorithm}[H] 
\caption{Single Measurement Inverse Solver}
\label{alg:patch_ss}
\begin{algorithmic}
\Require $\sigma_1 < \sigma_2 <  \ldots < \sigma_T$, $\epsilon > 0$, $P, M, \vy, K$
\State Initialize $\vx \sim \mathcal{N}(0, \sigma_T^2 \mI)$
\For{$t=T:1$}
\If{$t$ \text{mod} $K = 0$} 
\State Compute $\Dvtheta(\vx_t)$ using \eqref{eq:denoiser_stochastic} with 
\State \quad a random index $i$
\State Run $M$ iterations of CG initialized
\State \quad with $\Dvtheta(\vx_t)$ to obtain $\Dvtheta (\vx_t | \vy)$
\State Define $L(\vtheta) = \| \vy - \mA \Dvtheta (\vx_t | \vy)\|_2^2$
\State Update \vtheta by backpropagating $L(\vtheta)$
\EndIf
\State Sample $\vz \sim \mathcal{N}(0, \sigma_t^2 \mI)$
\State Set $\alpha_t = \epsilon \cdot \sigma_t^2$
\State Compute $D(\vx_t)$ using \eqref{eq:denoiser_stochastic} with a 
\State \quad random index $i$
\State Run $M$ iterations of CG for \eqref{eq:cg-step}
\State \quad initialized with $D(\vx_t)$
\State Set $\vs_t = (D - \vx_t)/\sigma_t^2$
\State Set $\vx_{t-1}$ to $\vx_t + \frac{\alpha_t}{2} \vs_t + \sqrt{\alpha_t} \vz$
\EndFor
\end{algorithmic}
\end{algorithm}
\end{minipage}
\end{wrapfigure}

Previously, additional LoRA parameters \citep{lora} have been used
as an injection to the network
to leave the original parameters unchanged during this process \citep{chung_scd, chung_ddip}.
However, the effect of using different ranks for LoRA
versus other methods of network fine-tuning on DIS has not been studied extensively,
so we opt to update all the weights of the network in this step.
Appendix \ref{ablation_tables} shows results from using the LoRA module. 

Crucially, iterative usage of CG for computing the conditional denoiser
allows for simple and efficient backpropagation through this loss function,
a task that would be much more computationally challenging
if another DIS such as \citet{chung:23:dps} or \citet{DDNM} were used.
Furthermore, because the number of diffusion steps is large
and the change in $\vx_t$ is small between consecutive timesteps,
we apply this network refining step 
only for certain iterations of the diffusion process,
reducing the computational burden.

After this step,
we apply the refined network to compute a new estimate of the score of $\vx_t$
and then use it to update $\vx_t$.
Similar to the network refining step,
we use the stochastic version of the denoiser given by \eqref{eq:denoiser_stochastic}
rather than the full version.
\citet{hu2024learningimagepriorspatchbased} showed that for patch-based priors,
Langevin dynamics \cite{song:19:gmb} works particularly well as a sampling algorithm,
so we use it here in conjunction with CG steps to enforce data fidelity.
Algorithm \ref{alg:patch_ss}
summarizes the entire method
for cases where only a single measurement \y is available.

\subsection{Small dataset setting}

Now turn to the case
where we have trained a diffusion model on OOD data,
but we also have a very small dataset of in-distribution test data
that we can use to fine-tune the model. 
When fine-tuning,
we initialize the network with the checkpoint trained on OOD data
and then use the denoising score matching loss function
to fine-tune the network on in-distribution data.
\citet{wang2023patch} found that improved image generation performance can be obtained
by training with varying patch sizes,
as opposed to fixing the patch size to the one used during inference.
Here, we apply a varying patch size scheme during fine-tuning also as a method of data augmentation.
We use the UNet architecture in \cite{ho:20:ddp}
that can accept images of different sizes.
Hence, the loss becomes
\begin{equation} \label{DSM_finetune}
\argmin_{\vtheta}
\mathbb{E}_{t \sim \mathcal{U}(0,T)} \mathbb{E}_{\vx \sim p_d(\vx)}
\mathbb{E}_{\rvepsilon \sim \mathcal{N} (0, \sigma_t^2 I)}
\| \Dvtheta(\vx + \rvepsilon, \sigma_t) - \vx \|_2^2,
\end{equation}
where $\vx \sim p_d(\vx)$ represents the drawing a randomly sized patch from an image belonging to the fine-tuning dataset.
Appendix \ref{sec:app_params} provides full details of the training process.

At reconstruction time, we assume that our network has been fine-tuned reasonably to our dataset.
Thus, we remove the network refining step in Algorithm \ref{alg:patch_ss}
and keep the weights fixed throughout the entire process.
We still use the same CG descent algorithm to enforce data fidelity with the measurement.

\section{Experiments}

\textbf{Experimental setup.}
For the CT experiments, we used the AAPM 2016 CT challenge data from \citet{AAPMct}.
We applied the same data processing methods as in \citet{hu2024learningimagepriorspatchbased}
with the exception that
we used all the XY slices from the 9 training volumes
to train the in distribution networks, yielding a total of approximately 5000 slices.
For the deblurring and superresolution experiments,
we used the CelebA-HQ dataset \citep{celeba} with each image having size $256 \times 256$.
The test data was a randomly selected subset of 10 of the images not used for training.
In all cases, we report the average metrics across the test images:
peak SNR (PSNR) in dB, and structural similarity metric (SSIM) \citep{ssim}.
For the training data,
we trained networks on generated phantom images consisting of
randomly placed ellipses of different shapes and sizes.
See Fig.~\ref{fig:side_by_side} for examples.
These phantoms can be generated on the fly in large quantities.
We used networks trained on grayscale phantoms for the CT experiments
and networks trained on RGB phantoms for the deblurring and superresolution experiments.
Appendix \ref{sec:phantoms_specs} contains precise specifications of the phantoms.

We trained the patch-based networks with $64 \times 64$ patches
and used a zero padding value of 64,
so that 5 patches in both directions were used to cover the target image.
We used the network architecture in \citet{karras2022elucidating}
for both the patch-based networks and whole-image networks.
All networks were trained on PyTorch using the Adam optimizer with 2 A40 GPUs. 

\textbf{Single measurement setting.}
In cases where no training data is available and we only have the measurement \y,
we applied Algorithm \ref{alg:patch_ss} to solve a variety of inverse problems:
CT reconstruction, deblurring, and superresolution. 
For the forward and backward projectors in CT reconstruction,
we used the implementation provided by the ODL \citet{ODL}. 
We performed two sparse-view CT (SVCT) experiments:
one using 20 projection views, and one using 60 projection views.
Both of these were done using a parallel beam forward projector
where the detector size was 512 pixels.
For the deblurring experiments,
we used a uniform blur kernel of size $9 \times 9$
and added white Gaussian noise with $\sigma=0.01$
where the clean image was scaled between 0 and 1.
For the superresolution experiments, we used a scaling factor of 4
with downsampling by averaging
and added white Gaussian noise with $\sigma=0.01$.

For the comparison methods,
we ran experiments that naively used the OOD diffusion model
without the self-supervised network refining process.
For reference, we also ran experiments using a diffusion model
trained on the entire in-distribution training set (the ``correct" model).
In practice, it would not be possible to obtain
such a large training dataset of in-distribution images.
Additionally, for these diffusion model methods,
we implemented both the patch-based version as well as the whole-image version.
The whole-image networks were trained with the loss function in \eqref{DSM}
and used the same network architecture as the patch-based models,
but the input of the network was the entire image
and did not contain positional encoding information.

We also compared with more traditional methods:
applying a simple baseline, reconstructing via the total variation regularizer (ADMM-TV),
and two plug and play (PnP) methods: PnP-ADMM \citep{xiaojian11} and PnP-RED \citep{xiaojian6}. 
For CT,
the baseline was obtained
by applying the filtered back-projection method to the measurement $\vy$.
For deblurring,
the baseline was simply equal to the blurred image.
For superresolution,
the baseline was obtained by upsampling the low resolution image
and using nearest neighbor interpolation.
The implementation of ADMM-TV can be found in \citet{Hong_2024}. 
Finally, since we assume we do not have access to any clean training data,
we used the off the shelf denoiser BM3D \citep{dabov2006bm3d}.
Appendix \ref{sec:app_params}
contains the values of all the parameters of the algorithms.

\begin{figure*}[ht!]
\centering
\includegraphics[width=0.99\linewidth]{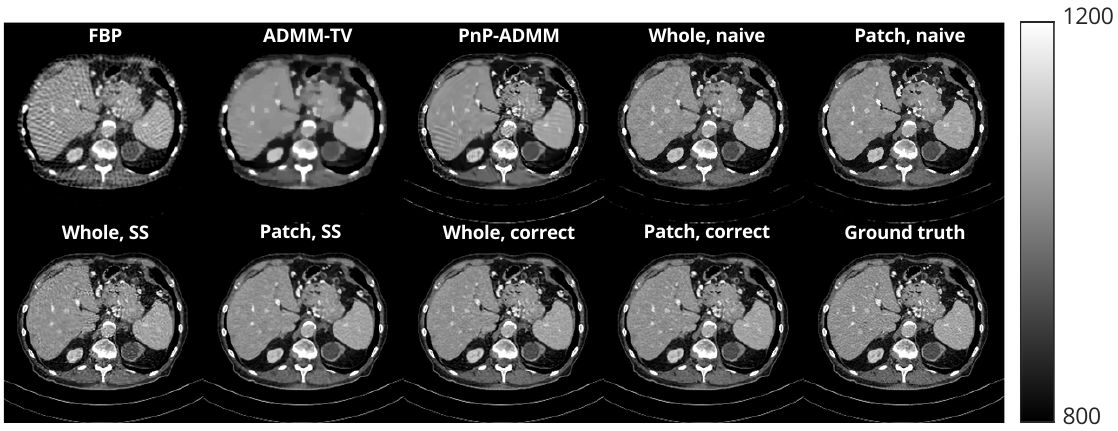}
\caption{Results of 60 view CT reconstruction using self supervised (SS) approach.
The display uses modified HU units
to show more contrast between organs.}
\label{fig: ct60_SS}
\end{figure*}

Table \ref{main_results} shows the main results
for single-measurement inverse problem solving.
The bottom two rows show the hypothetical performance
if it were possible to train a diffusion model on a large dataset of in distribution images,
which is not available in practice.
Our self-supervised patch-based diffusion approach
achieved significantly higher quantitative results
when averaged across the test dataset
than the self-supervised whole-image approach in all the inverse problems.
Furthermore, although the diffusion model that was initially used in this algorithm
was trained on completely different images, by applying the self-supervised loss,
the patch-based approach is able to achieve results that are close to
(and for the deblurring case, even surpassing) those using the in-distribution networks.
The table also shows that by including the self-supervised step,
a dramatic improvement over naively using the OOD model is achieved.
Lastly, Fig.~\ref{fig: ct60_SS} shows that some artifacts appear in the whole-image SS method 
that are not present in our patch SS method.


\setlength{\tabcolsep}{5pt}
\begin{table}[ht]
\centering
\begin{center}
\caption{Comparison of quantitative results on three different inverse problems in the single measurement setting.
Results are averages across all images in the test dataset.
Best results for practical use are in bold.}
\label{main_results}
\adjustbox{max width=\textwidth}{
\begin{tabular}{l|ll|ll|ll|ll}

\hline
\multirow{2}{*}{Method} &
\multicolumn{2}{c|}{CT, 20 Views} &
\multicolumn{2}{c|}{CT, 60 Views} &
\multicolumn{2}{c|}{Deblurring} &
\multicolumn{2}{c}{Superresolution}
\\
& PSNR$\uparrow$ & SSIM$\uparrow$
& PSNR$\uparrow$ & SSIM$\uparrow$
& PSNR$\uparrow$ & SSIM$\uparrow$
& PSNR$\uparrow$ & SSIM$\uparrow$
\\
\hline
Baseline
& 24.93 & 0.613 & 30.15 & 0.784 & 23.93 & 0.666 & 25.42 & 0.724 \\
ADMM-TV
& 26.81 & 0.750 & 31.14 & 0.862 & 27.58 & 0.773 & 25.22 & 0.729 \\
PnP-ADMM \citep{xiaojian11}
& 30.20 & 0.838 & 36.75 & 0.932 & 28.98 & 0.815 & 27.29 & 0.796 \\
PnP-RED \citep{xiaojian6}
& 27.12 & 0.682 & 32.68 & 0.876 & 28.37 & 0.793 & 27.73 & 0.809 \\
Whole image, naive
& 28.11 & 0.800 & 33.10 & 0.911 & 25.85 & 0.742 & 25.65 & 0.742 \\
Patches, naive \citep{hu2024learningimagepriorspatchbased}
& 27.44 & 0.719 & 33.97 & 0.934 & 26.77 & 0.782 & 26.12 & 0.759 \\
Self-supervised, whole  \citep{chung_scd}
& 33.19 & 0.861 & 40.47 & 0.957 & 29.50 & 0.831 & 27.07 & 0.701 \\
Self-supervised, patch (Ours)
& \textbf{33.77} & \textbf{0.874} & \textbf{41.45} & \textbf{0.969} & \textbf{30.34} & \textbf{0.860} & \textbf{28.10} & \textbf{0.827} \\
\hline 
Whole image, correct$^*$
& 33.99 & 0.886 & 41.67 & 0.969 & 29.87 & 0.851 & 28.33 & 0.801 \\
Patches, correct$^*$
& 34.02 & 0.889 & 41.70 & 0.967 & 30.12 & 0.865 & 28.49 & 0.835 \\
\hline
\end{tabular} 
}
\end{center}
{\raggedright *not available in practice for mismatched distribution inverse problems \par}
\vspace{-8pt}
\end{table}

To demonstrate that our method also works well
even when the mismatched distribution is closer to the true distribution,
we also ran an experiment
where the networks were initially trained on the LIDC-IDRI dataset \citep{lidc}.
We extracted 10000 2D slices from the 3D volumes
and rescaled all the images so that the pixel values were between 0 and 1.
We then ran Algorithm 1 to perform CT reconstruction
where the test dataset was the same as the one used in Table~\ref{main_results}.
Table~\ref{lidc_ss} shows the results of this experiment.
Our method achieved better quantitative results than the whole image method
and even outperformed the reconstructions using the in distribution network
but without any self-supervision.
Appendix~\ref{sec:inv_prob_figs} shows the visual results of these experiments. 
Appendix~\ref{sec:different_mismatch} further discusses using self-supervision
in cases where the initial network was trained on in-distribution data and shows improved image quality. 

We also ran ablation studies to examine the effect of various parameters on the proposed method.
\citet{chung_scd} and \citet{chung_ddip} used the LoRA module
for solving single-measurement inverse problems with diffusion models. 
We tested this method for CT reconstruction and deblurring with different rank adjustments
and found this method to be inferior to modifying the weights of the entire network.
We also ran experiments using networks with different numbers of weights.
Appendix~\ref{ablation_tables}
shows the results of these experiments.

\begin{figure*}[ht!]
\centering
\includegraphics[width=0.99\linewidth]{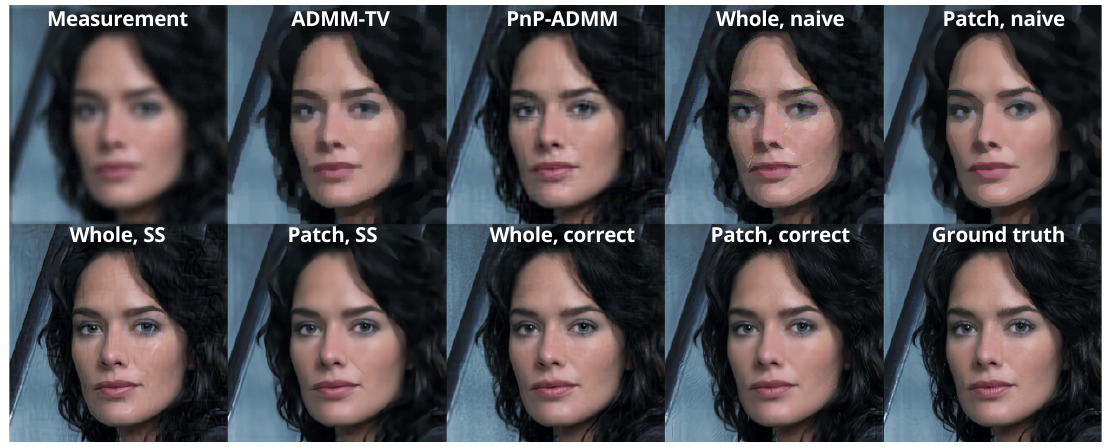}
\caption{Results of deblurring using self supervised (SS) approach
and comparison methods.}
\label{fig: deblur_ss}
\end{figure*}

\textbf{Small dataset setting.}
We ran experiments on the same inverse problems as the single measurement case.
The OOD networks were fine-tuned with 10 images randomly selected
from the in-distribution training set;
we also ran ablation studies using different quantities of in-distribution data
in Appendix \ref{ablation_tables}.
Figures \ref{fig:overfitting} and \ref{fig:overfitting_ssim} show that the patch-based model is much less prone to overfitting than the whole-image model. 
Hence, to evaluate the best possible performance of the whole-image model compared to the patch-based model, for both models
we chose the checkpoint yielding the best results for solving inverse problems.

\begin{figure*}[ht!]
\centering
\includegraphics[width=0.99\linewidth]{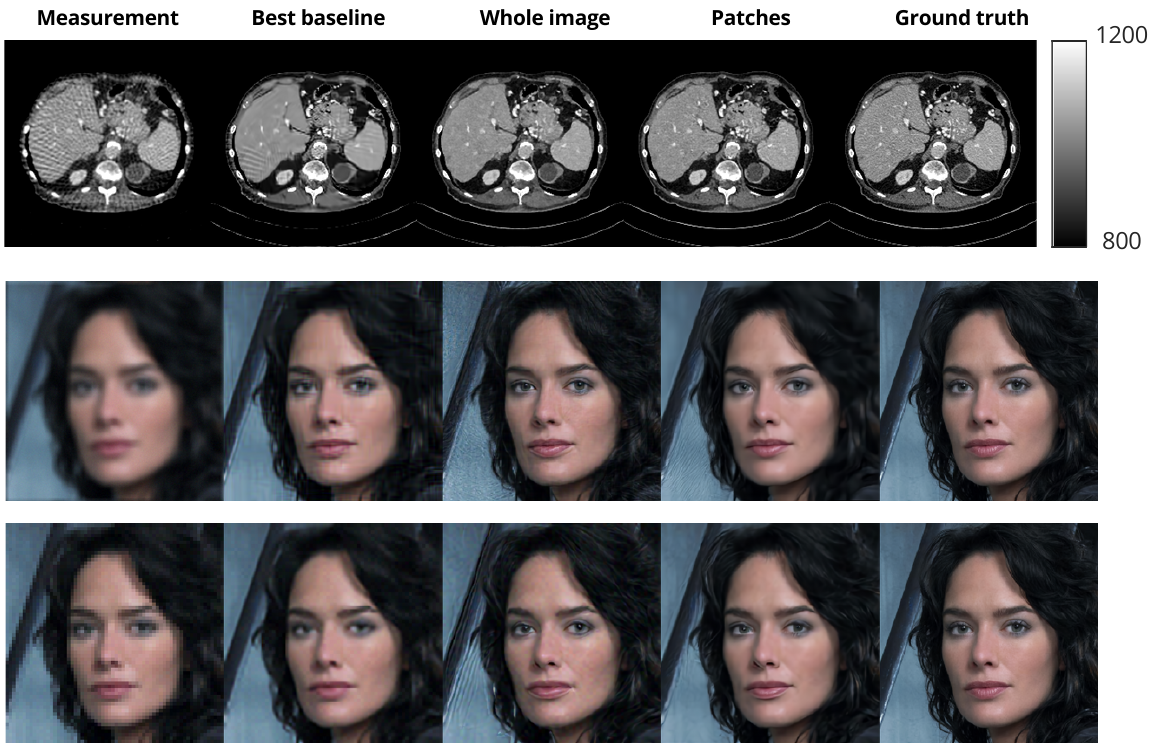}
\caption{Results of inverse problem solving in the small dataset setting. Top row is 60 view CT recon, middle row is deblurring, and bottom row is superresolution. For CT, measurement refers to FBP.}
\label{fig: finetune10}
\end{figure*}

Table \ref{ft10_results} shows the main results for solving inverse problems
using the fine-tuned diffusion model.
We compared the results of fine-tuning the whole-image model
with fine-tuning the patch-based model
as well as the best baseline out of the four baselines shown in Table \ref{main_results}.
The results show that the proposed patch-based method achieved the best performance
in terms of quantitative metrics for all of the inverse problems.
Figure \ref{fig: finetune10} shows the visual results of this experiment. 
The patch-based model is able to learn an acceptable prior
using the very small in-distribution dataset
and the reconstructed images contain fewer artifacts than the comparison methods.

\setlength{\tabcolsep}{5pt}
\begin{table}[ht]
\centering
\begin{center}
\caption{Comparison of results for using diffusion models fine-tuned
on 10 in-distribution images to solve inverse problems in small dataset setting.
Best results are in bold.}
\label{ft10_results}
\adjustbox{max width=\textwidth}{
\begin{tabular}{l|ll|ll|ll|ll}

\hline
\multirow{2}{*}{Method} &
\multicolumn{2}{c|}{CT, 20 Views} &
\multicolumn{2}{c|}{CT, 60 Views} &
\multicolumn{2}{c|}{Deblurring} &
\multicolumn{2}{c}{Superresolution}
\\
& PSNR$\uparrow$ & SSIM$\uparrow$
& PSNR$\uparrow$ & SSIM$\uparrow$
& PSNR$\uparrow$ & SSIM$\uparrow$
& PSNR$\uparrow$ & SSIM$\uparrow$
\\
\hline
Best baseline 
& 30.20 & 0.838 & 36.75 & 0.932 & 28.98 & 0.815 & 27.73 & 0.809 \\
Whole image
& 33.09 & \textbf{0.875} & 40.54 & 0.964 & 28.41 & 0.812 & 27.29 & 0.775 \\
Patches (Ours)
& \textbf{33.44} & \textbf{0.875} & \textbf{41.21} & \textbf{0.965} & \textbf{29.25} & \textbf{0.840} & \textbf{28.10} & \textbf{0.827} \\
\hline 
Patches, correct$^*$
& 34.02 & 0.889 & 41.70 & 0.967 & 30.12 & 0.865 & 28.49 & 0.835 \\
\hline
\end{tabular} 
}
\end{center}
{\raggedright *not available in practice for mismatched distribution inverse problems \par}
\vspace{-8pt}
\end{table}

Table \ref{overfitting}
further investigates the effect of overfitting.
For different amounts of training time using the small in-distribution dataset,
we ran the reconstruction algorithm for 60-view CT.
While the whole-image model exhibited substantial image degradation
when the network was fine-tuned for too long,
the patch-based model retained relatively stable performance throughout the entire training process. 
This illustrates that whole-image diffusion models exhibits severe overfitting problems
when only a small amount of training data is unavailable.
Furthermore, patch-based diffusion models assist greatly with this problem
and the results are evident for solving inverse problems.
Appendix \ref{sec:inv_prob_figs} shows the visual results of these experiments.

\begin{figure}[ht!]
  \centering
  \begin{minipage}{0.45\textwidth}
    \centering
    \includegraphics[width=\textwidth]{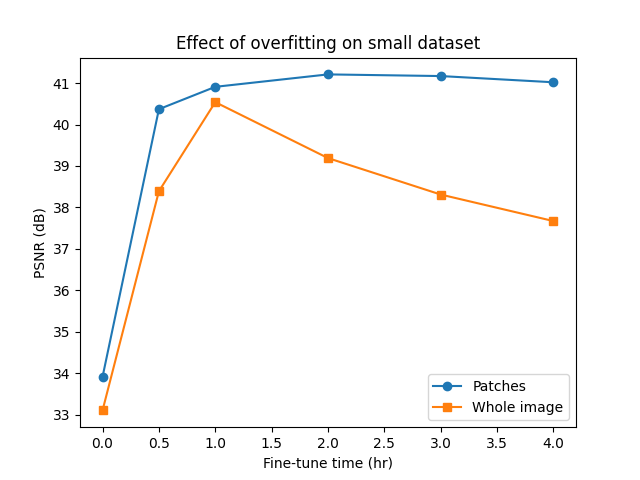}
    \caption{Comparison of PSNR between patch-based model and whole-image model for overfitting in small dataset setting.}
    \label{fig:overfitting}
  \end{minipage}
    \hspace{0.5cm}  
    \begin{minipage}{0.45\textwidth}
    \centering
    \includegraphics[width=\textwidth]{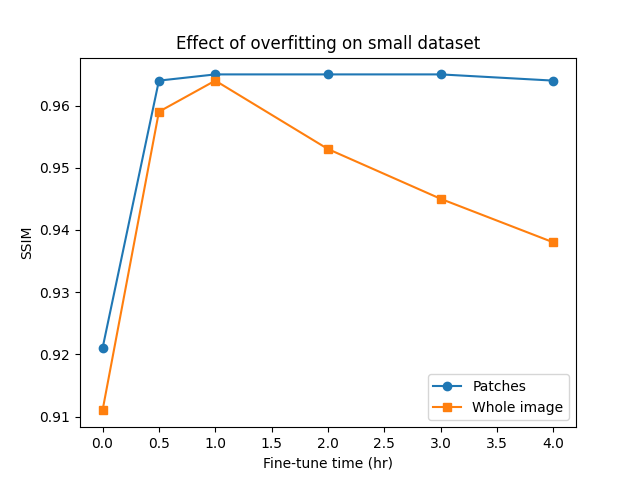}
    \caption{Comparison of SSIM between patch-based model and whole-image model for overfitting in small dataset setting.}
    \label{fig:overfitting_ssim}
  \end{minipage}
    
\end{figure}

\begin{figure*}[ht!]
\centering
\includegraphics[width=0.99\linewidth]{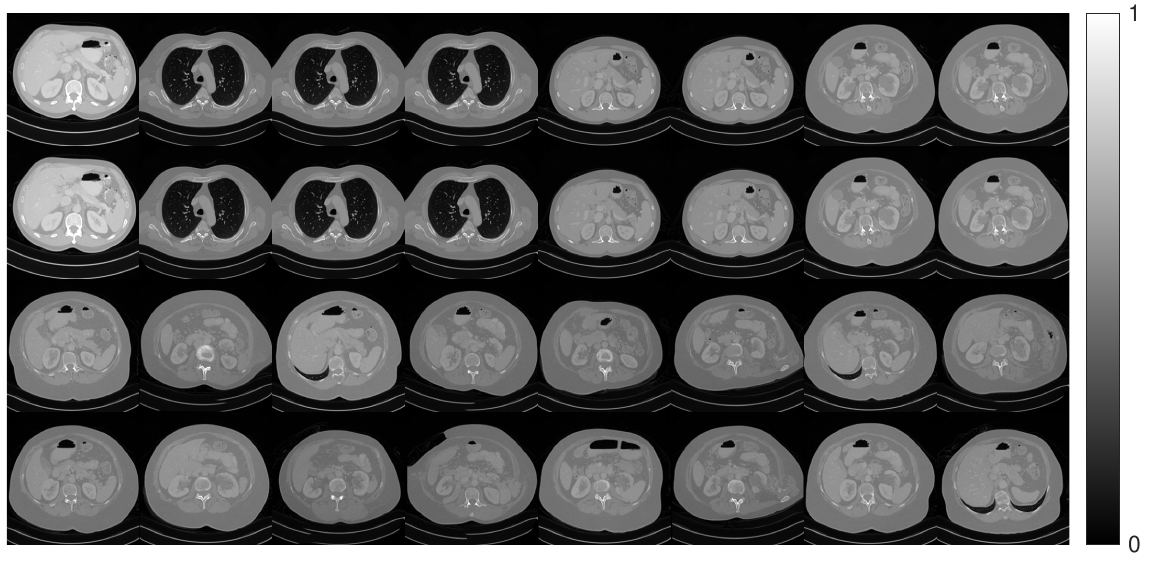}
\caption{Unconditional generation of CT images from networks fine-tuned in the small dataset setting.
Top two rows were generated with the whole image model;
bottom two rows were generated with the patch-based model.}
\label{fig: generation}
\end{figure*}

To look at the priors learned by the different models from fine-tuning,
we unconditionally generated images from the checkpoints obtained by fine-tuning
on the 10 image CT dataset.
Figure \ref{fig: generation} shows a subset of the generated images
where we used the checkpoints obtained after 4 hours of training.
The top two rows consist of images generated by the whole-image model
and the bottom two rows consist of images generated by the patch diffusion model. 
To emphasize the memorization point,
we grouped together similar looking images in the top two rows:
it can be seen that the images in each group look virtually identical,
despite the fact that the pure white noise initializations for each sample was different. 
On the other hand, while the samples generated by the patch diffusion model
also show some unrealistic features,
they all show some distinct features,
which implies that this model has much better generalization ability.

\section{Conclusion}
This paper presented a method of using patch-based diffusion models
to solve inverse problems when the data distribution is mismatched from the trained network.
In particular, we conducted experiments in the setting when only a single measurement is available as well as the setting when a very small subset of in-distribution data is available.
In both settings, the proposed patch-based method outperformd whole-image methods
in a variety of inverse problems.
In the future, more work could be done on using acceleration methods for faster reconstruction,
exploring other less computationally expensive methods of fine-tuning the network
geared toward inverse problem solving,
and methods of refining the prior when a set of measurements are available
\citep{yaman:20:ssl}.
Limitations of the work include a slow runtime for the self-supervised algorithm
and a lack of theoretical guarantees for convergence of algorithms and dataset size requirements.




\subsection*{Acknowledgments}


\bibliography{iclr2025_conference}
\bibliographystyle{iclr2025_conference}

\appendix

\clearpage
\section{Appendix}

\subsection{Additional inverse problem solving figures}
\label{sec:inv_prob_figs}

\begin{figure*}[ht!]
\centering
\includegraphics[width=0.99\linewidth]{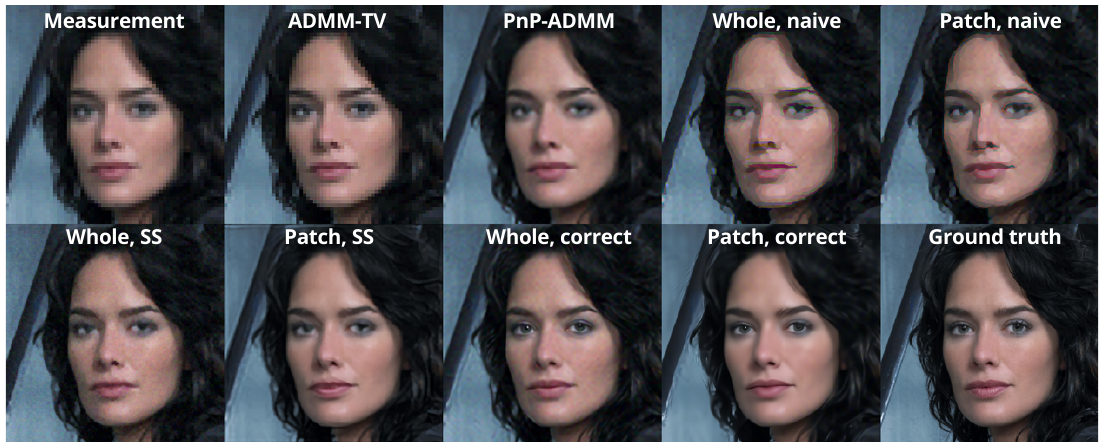}
\caption{Results of superresolution using self supervised (SS) approach
and comparison methods.}
\label{fig: super_ss}
\end{figure*}

Figure \ref{fig: super_ss} shows the results of various methods applied to superresolution in the single measurement setting.

\begin{figure*}[ht!]
\centering
\includegraphics[width=0.99\linewidth]{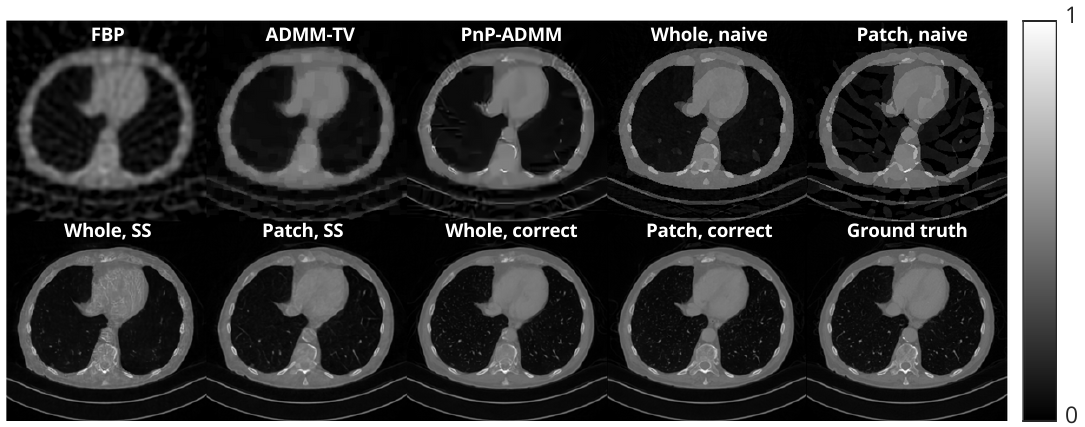}
\caption{Results of 20 view CT reconstruction in a self-supervised setting.
For clarity, the images are plotted on the same scale as the diffusion models were trained.}
\label{fig: 20view_SS}
\end{figure*}

Figure \ref{fig: 20view_SS} shows the results of 20 view CT reconstruction using Algorithm \ref{alg:patch_ss}.
This very sparse view CT recon problem is made more challenging by the lack of any training data.
Artifacts can clearly be seen in all the comparison methods.
Despite this challenge, reconstructions such as this one can still be useful for medical applications such as patient positioning.

\begin{figure*}[ht!]
\centering
\includegraphics[width=0.99\linewidth]{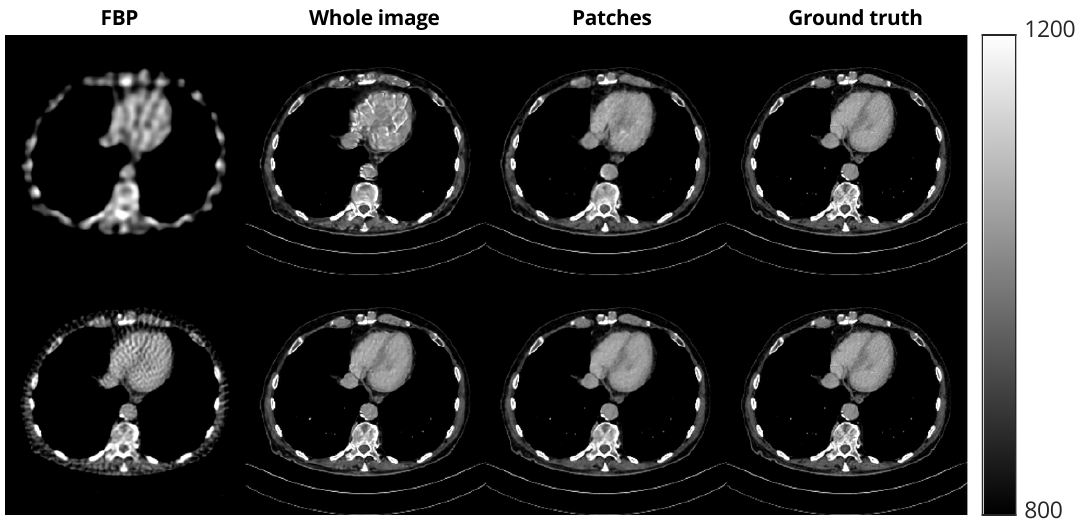}
\caption{Results of CT reconstruction in a self-supervised setting
when the starting network was trained on the LIDC dataset.
Top row used 20 views and bottom row used 60 views.}
\label{fig: ablation4}
\end{figure*}

Figure \ref{fig: ablation4} shows the results of running self-supervised CT reconstruction with 20 views and 60 views
where the starting checkpoint was obtained through training on a large (but out of distribution) CT dataset:
10000 LIDC-IDRI slices.
Particularly for 20 views, the artifacts from using the whole image model are apparent,
while the patch-based model obtains a much higher quality reconstruction.
Thus, regardless of whether the starting network has a severely mismatched distribution (ellipses)
or a slightly mismatched distribution (different CT dataset), our proposed method outperforms the whole image model.

\begin{figure*}[ht!]
\centering
\includegraphics[width=0.99\linewidth]{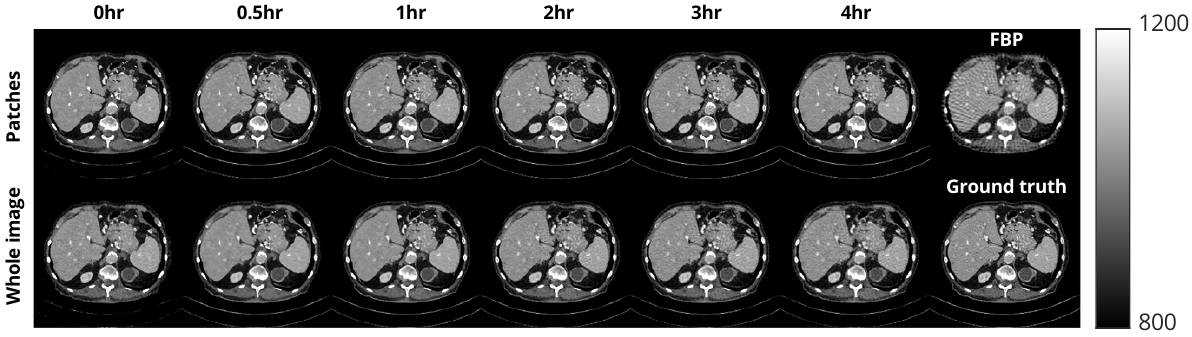}
\caption{Results of 60 view CT recon with networks fine-tuned on 10 in distribution CT images
for varying amounts of training time.}
\label{fig: ablation2}
\end{figure*}

Figure \ref{fig: ablation2} shows the results of performing 60 view CT reconstruction
in an unsupervised manner
from checkpoints fine-tuned using the small in distribution CT dataset.
The images on the bottom row shows the progressively worsening degradation
and increasing number of artifacts
resulting from overfitting exhibited by whole image model.
On the other hand, the top row shows relatively stable performance exhibited by the patch-based model
as it is able to avoid overfitting much better.

\setlength{\tabcolsep}{5pt}
\begin{table}[ht]
\centering
\begin{center}
\caption{Performance of fine-tuning on 60 view CT using checkpoints
trained for different lengths of time. Best results are in bold.}
\label{overfitting}
\begin{tabular}{c|cc|cc}
 \toprule
 Train & \multicolumn{2}{c|}{Patches} & \multicolumn{2}{c}{Whole image}\\
 time (hr) & PSNR$\uparrow$ & SSIM $\uparrow$ & PSNR$\uparrow$ & SSIM $\uparrow$\\
 \midrule
  0 & 33.91 & 0.921 & 33.10 & 0.911 \\
  0.5 & 40.37 & 0.964 & 38.39 & 0.959 \\
  1 & 40.91 & \textbf{0.965} & \textbf{40.54} & \textbf{0.964} \\
 2 & \textbf{41.21} & \textbf{0.965} & 39.19 & 0.953 \\
 3 & 41.17 & \textbf{0.965} & 38.31 & 0.945 \\
 4 & 41.02 & 0.964 & 37.67 & 0.938 \\
 \bottomrule
\end{tabular}
\end{center}
\end{table}

\subsection{Effect of self-supervision for different distributions}
\label{sec:different_mismatch}
\begin{figure}[ht!]
  \centering
  \begin{minipage}{0.49\textwidth}
    \centering
    \includegraphics[width=\textwidth]{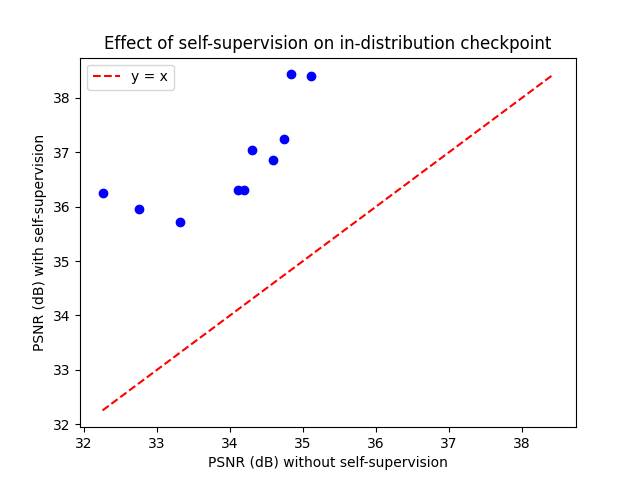}
    \caption{PSNR of 20 view CT reconstruction in single-measurement setting
    using a patch-based in-distribution network.}
    \label{fig:ablation9_20view}
  \end{minipage}
    \hspace{0.1cm}  
    \begin{minipage}{0.49\textwidth}
    \centering
    \includegraphics[width=\textwidth]{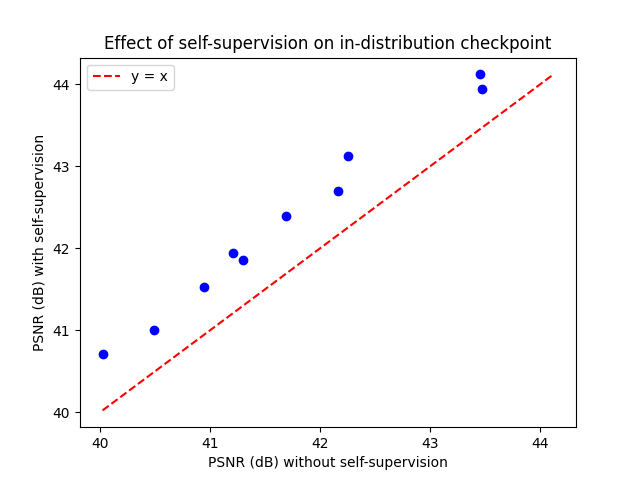}
    \caption{PSNR of 60 view CT reconstruction in single-measurement setting
    using a patch-based in-distribution network.}
    \label{fig:ablation9_60view}
  \end{minipage}
\end{figure}

Recall that in the single measurement setting,
Algorithm \ref{alg:patch_ss} is used to adjust the underlying distribution of the network
away from the originally trained OOD data and toward the ground truth image.
We investigated the effect of applying this method
even when the network was trained on the in-distribution data.
Figures \ref{fig:ablation9_20view} and \ref{fig:ablation9_60view}
show the results of this experiment for CT reconstruction,
where each point represents the specific PSNR for one of the images in the test dataset. 
If the additional self-supervision step had no effect on the image quality,
the points would lie on the red line. 
However, in both cases, all of the points are above the red line,
indicating that the self-supervision step of the algorithm
improves the image quality even when the network was already trained on in-distribution data. 
Furthermore, the improvement is more substantial for for the 20 view case than the 60 view case,
as the predicted clean images $D_\theta(\vx_t | \vy)$ at each step for the 60 view case
are likely to be more closely aligned with the measurement,
so the network refining step becomes less significant.
Importantly, this shows that in practice,
one may directly apply Algorithm \ref{alg:patch_ss} to solve inverse problems
without knowledge of the severity of the mismatch in distribution between training and testing data:
even when there is no mismatch, the additional self-supervision step does not degrade the image quality.

Table \ref{ablation9} summarizes the results of using different training datasets
while keeping the same test dataset (AAPM CT images). 
The distribution shift is greatest when the network is trained on ellipse phantoms
and used to reconstruct the AAPM CT images, so the reconstruction quality is the lowest in this case. 
The LIDC dataset consists of CT images which belong to a distribution
that is reasonably similar to the distribution of AAPM CT images,
so when using the network trained on LIDC images,
the quality drop over using an in-distribution network is not substantial. 
Finally, the improvements obtained by using more in-distribution networks
is more apparent for the 20 view case as the measurements are sparser for this case,
so the prior plays a larger role in obtaining an accurate reconstruction.

\begin{table}[ht]
\centering
\begin{center}
\caption{Single measurement CT reconstruction results
where the initial checkpoint was trained on LIDC dataset
and refined on the fly with the AAPM measurement.
}
\label{lidc_ss}
\begin{tabular}{c|cc|cc}
 \toprule
 Dataset & \multicolumn{2}{c|}{CT, 20 views} & \multicolumn{2}{c}{CT, 60 views}\\
 size  & PSNR$\uparrow$ & SSIM $\uparrow$ & PSNR$\uparrow$ & SSIM $\uparrow$\\
 \midrule
  Whole image & 35.01 & 0.894 & 41.95 & 0.967 \\
  Patches (Ours) & 36.34 & 0.918 & 42.32 & 0.972 \\
 \bottomrule
\end{tabular}
\end{center}
\end{table}

\setlength{\tabcolsep}{5pt}
\begin{table}[ht]
\centering
\begin{center}
\caption{Performance of patch-based model in single measurement setting for CT reconstruction for different OOD training datasets.}
\label{ablation9}
\begin{tabular}{c|cc|cc}
 \toprule
 Train & \multicolumn{2}{c|}{20 views} & \multicolumn{2}{c}{60 views}\\
 time (hr) & PSNR$\uparrow$ & SSIM $\uparrow$ & PSNR$\uparrow$ & SSIM $\uparrow$\\
 \midrule
  Ellipses & 33.77 & 0.874 & 41.45 & 0.966 \\
  LIDC & 36.34 & 0.918 & 42.32 & 0.970 \\
  AAPM & 36.82 & 0.923 & 42.33 & 0.970 \\
 \bottomrule
\end{tabular}
\end{center}
\end{table}

\newpage 
\subsection{Ablation studies} \label{ablation_tables}
We performed four ablation studies
to evaluate the impact of various parameters on the proposed methods.
Similar to the main text,
all quantitative results are averaged across the test dataset.

\textbf{Low rank adaptation.}
To avoid overfitting to the measurement in self-supervised settings,
\citet{chung_scd} proposed using a low rank adaptation to the weights of the neural network,
reducing the number of weights that are adjusted during reconstruction by a factor of around 100.
Here we investigate the effect of using different ranks of adaptations on two inverse problems:
60 view CT reconstruction and deblurring.
Consistent with \citet{chung_scd} and \citet{chung_ddip},
we only used the LoRA module for attention and convolution layers.
We also allowed the biases of the network to be changed. 

\begin{figure*}[ht!]
\centering
\includegraphics[width=0.99\linewidth]{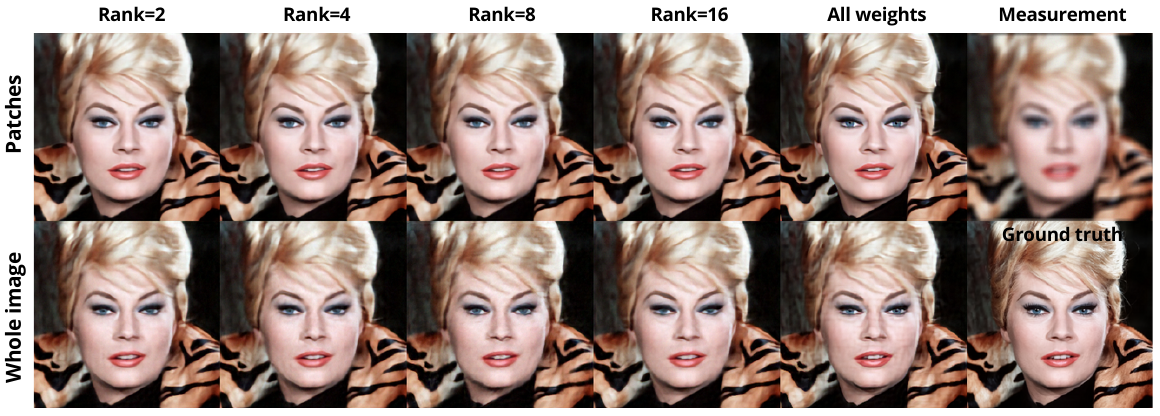}
\caption{Results of using LoRA module for deblurring in a single measurement setting.
All weights refers to adjusting all the weights of the network at reconstruction time.}
\label{fig: ablation5}
\end{figure*}

Tables \ref{ct_lora} and \ref{deblur_lora} show the quantitative results of these experiments,
where a rank of ``full" represents fine-tuning all the weights of the network. 
In all cases, using LoRA for this fine-tuning process results in worse reconstructions than simply fine-tuning the entire network.
The visual results are especially apparent in Figure \ref{fig: ablation5}:
the reconstructed image becomes oversmoothed when using LoRA and artifacts become present when using the whole image model.
This is likely due to the large distribution shift between the initial distribution of images and target distribution of faces:
the low rank adaptation to the mismatched network is not sufficient to represent the new distribution
and thus the self-supervised loss function results in smoothed images.

\begin{table}[ht]
\centering
\begin{center}
\caption{Performance of 60 view CT recon using self-supervised network refining with LoRA module. Best results are in bold.}
\label{ct_lora}
\begin{tabular}{cc|cc|cc}
 \toprule
 Rank & Parameters (\%) & \multicolumn{2}{c|}{Patches} & \multicolumn{2}{c}{Whole image}\\
 & & PSNR$\uparrow$ & SSIM $\uparrow$ & PSNR$\uparrow$ & SSIM $\uparrow$\\
 \midrule
  2 & 1.1 & 40.37 & 0.963 & 39.25 & 0.952 \\
  4 & 2.0 & 40.32 & 0.963 & 39.10 & 0.951 \\
  8 & 3.8 & 40.33 & 0.963 & 39.18 & 0.951 \\
 16 & 7.2 & 40.32 & 0.963 & 39.33 & 0.953 \\
 Full & 100 & \textbf{41.45} & \textbf{0.966} & \textbf{40.47} & \textbf{0.957} \\
 \bottomrule
\end{tabular}
\end{center}
\end{table}

\begin{table}[ht]
\centering
\begin{center}
\caption{Performance of deblurring using self-supervised network refining with LoRA module.
Best results are in bold.}
\label{deblur_lora}
\begin{tabular}{cc|cc|cc}
 \toprule
 Rank & Parameters (\%) & \multicolumn{2}{c|}{Patches} & \multicolumn{2}{c}{Whole image}\\
 & & PSNR$\uparrow$ & SSIM $\uparrow$ & PSNR$\uparrow$ & SSIM $\uparrow$\\
 \midrule
  2 & 1.1 & 29.31 & 0.830 & 29.19 & 0.811 \\
  4 & 2.0 & 29.31 & 0.829 & 29.35 & 0.817 \\
  8 & 3.8 & 29.38 & 0.831 & 29.19 & 0.810 \\
 16 & 7.2 & 29.31 & 0.830 & 29.33 & 0.815 \\
 Full & 100 & \textbf{30.34} & \textbf{0.860} & \textbf{29.50} & \textbf{0.831} \\
 \bottomrule
\end{tabular}
\end{center}
\end{table}

\begin{figure*}[ht!]
\centering
\includegraphics[width=0.99\linewidth]{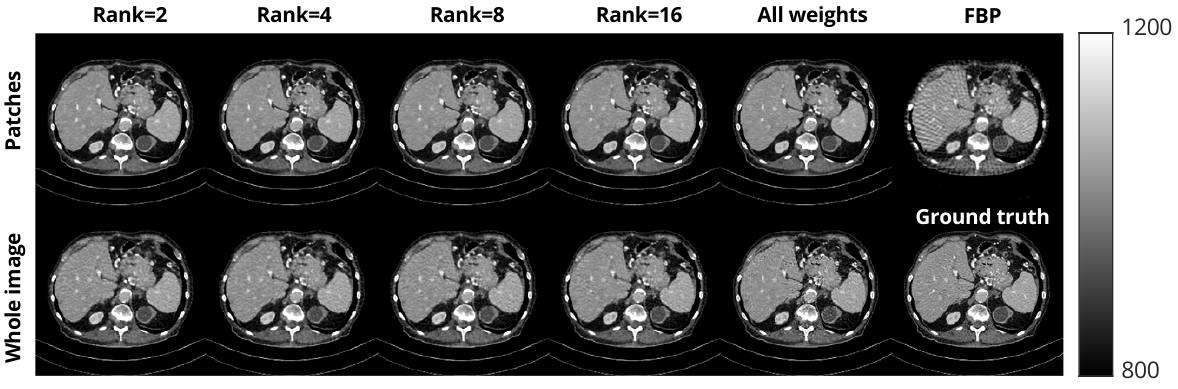}
\caption{Results of using LoRA module for 60 view CT reconstruction in a single measurement setting.
All weights refers to adjusting all the weights of the network at reconstruction time.}
\label{fig: ablation1}
\end{figure*}

\begin{figure}[ht!]
\centering
\includegraphics[width=0.99\linewidth]{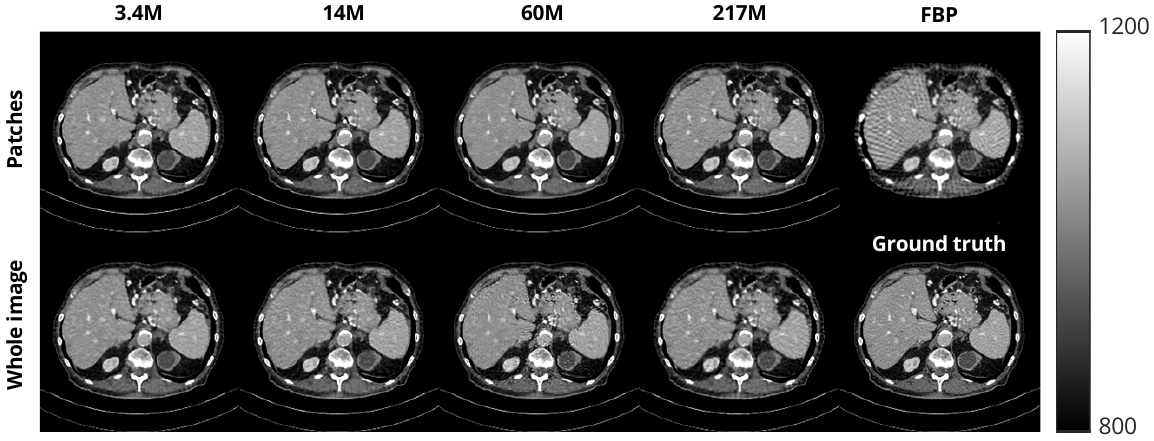}
\caption{Results of 60 view CT recon using networks with different numbers of parameters
in the single-measurement setting.
The top numbers show the number of total parameters in the network.}
\label{fig: ablation3}
\end{figure}

\textbf{Effect of network size.}
In the self-supervised case, another potential method to avoid overfitting
is to use a smaller network.
We trained networks with differing numbers of base channels (but no other modifications)
on the ellipse phantom dataset
and then used Algorithm \ref{alg:patch_ss} to perform self-supervised 60 view CT reconstruction.
Table \ref{ablation3} shows the quantitative results of this experiment.
For both the patch-based model and the whole image model,
the network with 128 base channels obtained the best result,
so we used this network architecture for all the main experiments.
Figure~\ref{fig: ablation3}again shows evidence of overfitting in the form of artifacts
in the otherwise smooth regions of the organs when using the network with 256 base channels.
These artifacts are less obvious in the patch-based model.

\begin{table}[ht]
\centering
\begin{center}
\caption{Performance of 60 view CT recon in a self-supervised manner with networks of different sizes.
Best results are in bold.}
\label{ablation3}
\begin{tabular}{cc|cc|cc}
 \toprule
 Base & Parameters & \multicolumn{2}{c|}{Patches} & \multicolumn{2}{c}{Whole image}\\
 Channels & (Millions) & PSNR$\uparrow$ & SSIM $\uparrow$ & PSNR$\uparrow$ & SSIM $\uparrow$\\
 \midrule
  32 & 3.4 & 39.73 & 0.958 & 39.69 & 0.957 \\
  64 & 14 & 40.37 & 0.961 & 40.07 & \textbf{0.958} \\
  128 & 60 & \textbf{41.45} & \textbf{0.966} & \textbf{40.47} & 0.957 \\
 256 & 217 & 40.29 & 0.959 & 39.28 & 0.954 \\
 \bottomrule
\end{tabular}
\end{center}
\end{table}

\textbf{Fine-tuning with a larger dataset.}
To examine the effect of fine-tuning the networks on differing sizes of in-distribution datasets,
we started with the same checkpoint trained on ellipses
and fine-tuned them using various sizes of datasets consisting of CT images.
Each small dataset consisted of randomly selected images from the entire 5000 image AAPM dataset.
Next we used these checkpoints to perform 60 view CT reconstruction (without any self supervision).
Table \ref{ablation6} shows the results of these experiments,
where we also included the results of using the in-distribution network
trained on the entire 5000 image dataset.
This shows that for a wide range of different fine-tuning dataset sizes
our proposed method obtained better metrics than the whole-image model.

\begin{table}[ht]
\centering
\begin{center}
\caption{Performance of fine-tuning on 60 view CT using checkpoints fine-tuned from different dataset sizes.
Best results are in bold.}
\label{ablation6}
\begin{tabular}{c|cc|cc}
 \toprule
 Dataset & \multicolumn{2}{c|}{Patches} & \multicolumn{2}{c}{Whole image}\\
 size  & PSNR$\uparrow$ & SSIM $\uparrow$ & PSNR$\uparrow$ & SSIM $\uparrow$\\
 \midrule
  3 & 40.93 & 0.964 & 40.45 & 0.964 \\
  10 & 41.21 & 0.965 & 40.54 & 0.964 \\
  30 & 41.31 & 0.966 & 40.66 & 0.967 \\
 100 & 41.46 & 0.967 & 40.96 & 0.968 \\
 5000* & \textbf{41.70} & \textbf{0.967} & \textbf{41.67} & \textbf{0.969} \\
 \bottomrule
\end{tabular}
\end{center}
\end{table}

\begin{figure*}[ht]
\centering
\includegraphics[width=0.99\linewidth]{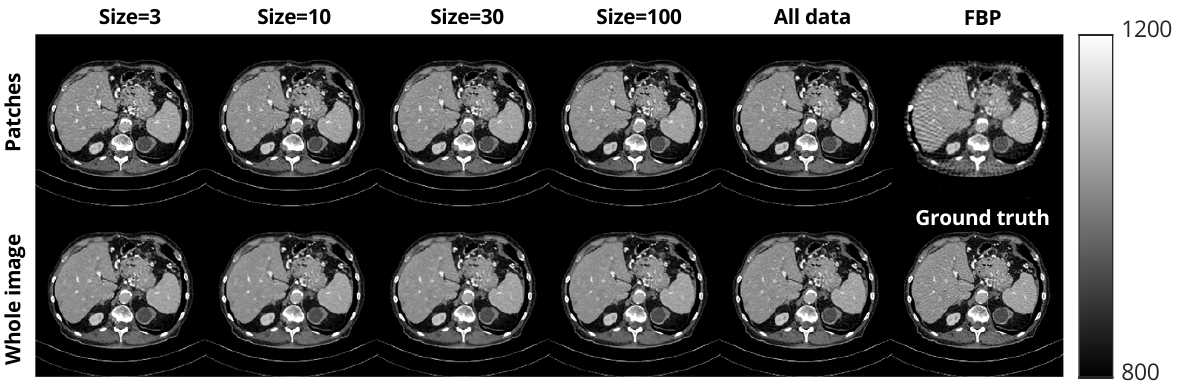}
\caption{Results of 60 view CT recon in the small dataset setting where the size of the small dataset is varied.}
\label{fig: ablation6}
\end{figure*}

\begin{figure}[ht!]
  \centering
  \begin{minipage}{0.45\textwidth}
    \centering
    \includegraphics[width=\textwidth]{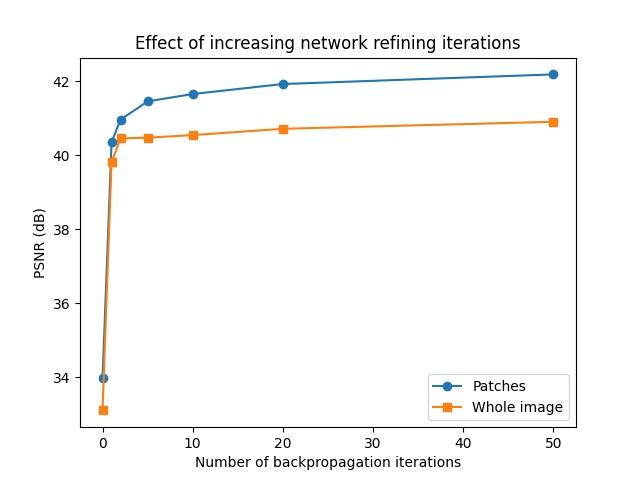}
    \caption{Comparison of PSNR between patch-based model and whole-image model for number of network refining iterations in single measurement setting.}
    \label{fig:ablation7_psnr}
  \end{minipage}
    \hspace{0.5cm}  
    \begin{minipage}{0.45\textwidth}
    \centering
    \includegraphics[width=\textwidth]{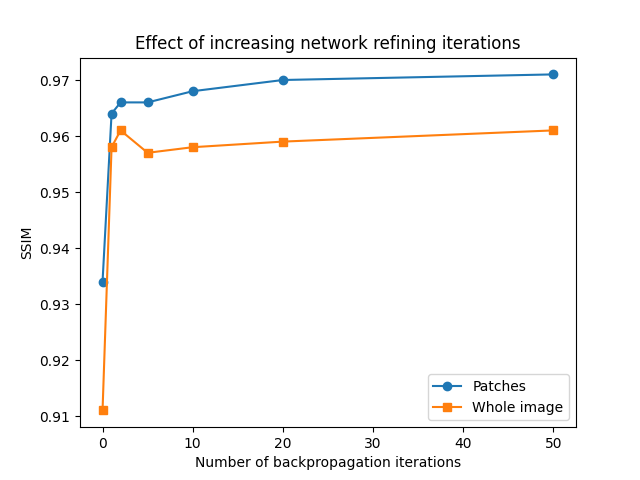}
    \caption{Comparison of SSIM between patch-based model and whole-image model for number of network refining iterations in single measurement setting.}
    \label{fig:ablation7_ssim}
  \end{minipage}
\end{figure}

\textbf{Backpropagation iterations during self-supervision.}
In the single measurement setting, the self-supervised loss is crucial
to ensuring that the OOD network output is consistent with the measurement. 
Backpropagation through the network is necessary to minimize this loss,
but too much network refining during this step could lead to overfitting
to the measurement and image degradation.
We ran experiments examining the effect of the number of backpropagation iterations
during each step for the patch-based model and the whole image model. 
Figures \ref{fig:ablation7_psnr} and \ref{fig:ablation7_ssim} show that in both cases,
performance generally improved when increasing the number of backpropagation iterations
and overfitting is avoided. 
Additionally, the patch-based model always outperformed the whole image model
and exhibited more improvement as the number of backpropagation iterations increased. 
For our main experiments,
we used 5 iterations as the improved performance became marginal compared to the extra runtime.

\begin{table}[ht]
\centering
\begin{center}
\caption{Performance of Algorithm \ref{alg:patch_ss} for 60 view CT reconstruction in single measurement setting with different numbers of backpropagation iterations. Best results are in bold.}
\label{ablation7}
\begin{tabular}{c|cc|cc}
 \toprule
 Backprop & \multicolumn{2}{c|}{Patches} & \multicolumn{2}{c}{Whole image}\\
 iterations  & PSNR$\uparrow$ & SSIM $\uparrow$ & PSNR$\uparrow$ & SSIM $\uparrow$\\
 \midrule
  0 & 33.97 & 0.934 & 33.10 & 0.911 \\
  1 & 40.35 & 0.964 & 39.81 & 0.958 \\
  2 & 40.96 & 0.966 & 40.45 & 0.961 \\
 5 & 41.45 & 0.966 & 40.47 & 0.957 \\
  10 & 41.65 & 0.968 & 40.54 & 0.958 \\
   20 & 41.92 & 0.970 & 40.71 & 0.959 \\
 50 & \textbf{42.18} & \textbf{0.971} & \textbf{40.90} & \textbf{0.961} \\
 \bottomrule
\end{tabular}
\end{center}
\end{table}

\newpage
\subsection{Phantom dataset details}
\label{sec:phantoms_specs}
\begin{figure}[htbp]
    \centering
    \begin{subfigure}{0.45\textwidth}
        \centering
        \includegraphics[width=\textwidth]{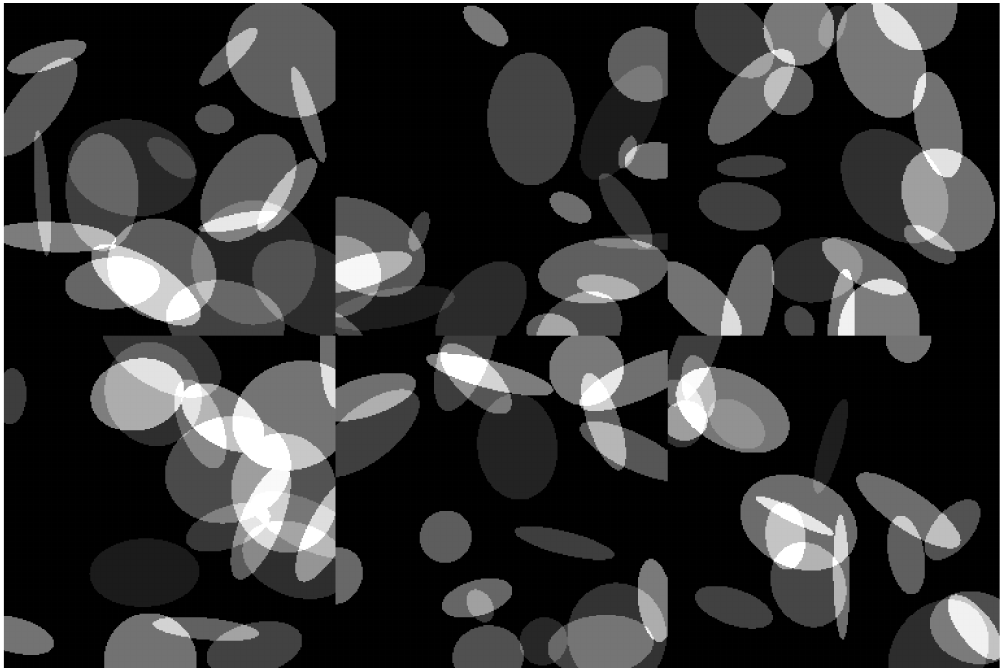}
        \caption{Six grayscale phantoms}
        \label{fig:1}
    \end{subfigure}
    \hfill
    \begin{subfigure}{0.45\textwidth}
        \centering
        \includegraphics[width=\textwidth]{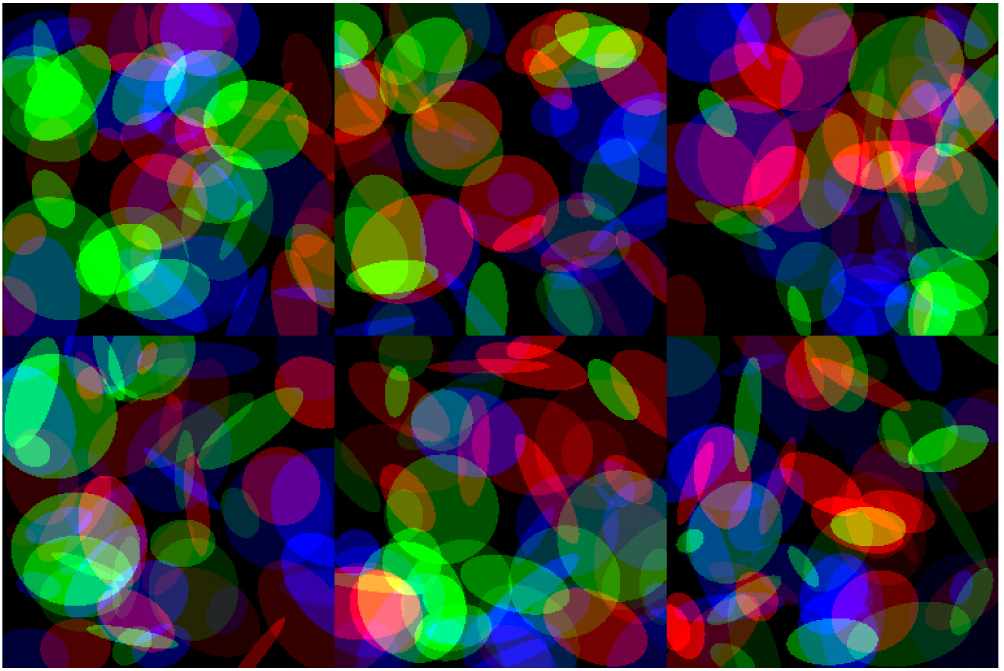}
        \caption{Six colored phantoms}
        \label{fig:2}
    \end{subfigure}
    \caption{Six sample grayscale phantoms and colored phantoms used to train the mismatched distribution diffusion models}
    \label{fig:side_by_side}
\end{figure}

We used two phantom datasets of 10000 images each:
one consisting of grayscale phantoms and the other consisting of colored phantoms.
The grayscale phantoms consisted of 20 ellipses with a random center within the image,
each with minor and major axis having length equal to a random number
chosen between 2 and 20 percent of the width of the image.
The grayscale value of each ellipse was randomly chosen between 0.1 and 0.5;
if two or more ellipses overlapped,
the grayscale values were summed for the overlapped area with all values exceeding 1 set to 1.
Finally, all ellipses were set to a random angle of rotation.
The colored phantoms were generated in the same way,
except the RGB values for each ellipse were set independently
and then multiplied by 255 at the end.
Figure \ref{fig:side_by_side} shows some of the sample phantoms.

\subsection{Experiment parameters}
\label{sec:app_params}
We applied the framework of \citet{karras2022elucidating} to train the patch-based networks and whole image networks.
Since images were scaled between 0 and 1 for both grayscale images and RGB channels,
we chose a maximum noise level of $\sigma=40$
and a minimum noise level of $\sigma=0.002$ for training.
We used the same UNet architecture for all the networks
consisting of a base channel multiplier size of 128
and 2, 2, and 2 channels per resolution for the three layers.
We also used dropout connections with a probability of 0.05
and exponential moving average for weight decay
with a half life of 500K images to avoid overfitting.

The learning rate was chosen to be $2 \cdot 10^{-4}$ when training networks from scratch
and was $1 \cdot 10^{-4}$ for the fine-tuning experiments.
For the patch-based networks, the batch size for the main patch size ($64 \times 64$) was 128,
although batch sizes of 256 and 512 were used for the two smaller patch sizes of $32 \times 32$ and $16 \times 16$. The probabilities of using these three patch sizes were $0.5, 0.3$, and $0.2$ respectively.
For the whole image model, we kept all the parameters the same,
but used a batch size of 8.

For image generation and inverse problem solving,
we used a geometrically spaced descending noise level
that was fine tuned to optimize the performance for each type of problem.
We used the same set of parameters
for the patch-based model and whole image model. 
The values without the self-supervised loss are as follows: 
\begin{itemize}
    \item CT with 20 and 60 views: $\sigma_{\max} = 10, \sigma_{\min} = 0.005$
    \item Deblurring: $\sigma_{\max} = 40, \sigma_{\min} = 0.005$
    \item Superresolution: $\sigma_{\max} = 40, \sigma_{\min} = 0.01$.
\end{itemize}
The values with the self-supervised loss are as follows: 
\begin{itemize}
    \item CT with 20 and 60 views: $\sigma_{\max} = 10, \sigma_{\min} = 0.01$
    \item Deblurring: $\sigma_{\max} = 1, \sigma_{\min} = 0.01$
    \item Superresolution: $\sigma_{\max} = 1, \sigma_{\min} = 0.01$.
\end{itemize}
Finally, for generating the CT images we used $\sigma_{\max} = 40, \sigma_{\min} = 0.005$.

When running Algorithm \ref{alg:patch_ss},
we set $K=10$ for all experiments and $M=5$ for CT reconstruction
and $M=1$ for deblurring and superresolution.
We ran 5 iterations of network backpropagation with a learning rate of $10^{-5}$.
When using the LoRA module as in the ablation studies (see Tables \ref{deblur_lora} and \ref{ct_lora}),
we ran 10 iterations of network backpropagation with a learning rate of $10^{-3}$.

The ADMM-TV method for linear inverse problems consists of solving the optimization problem 
\begin{equation}
    \text{argmax}_{\vx} \frac{1}{2} \| \vy - A \vx \|_2^2 + \lambda \, \text{TV}(\vx),
\end{equation}
where $\text{TV}(\vx)$ represents the L1 norm total variation of $vx$,
and the problem is solved with the alternating direction method of multipliers.
For CT reconstruction, deblurring, and superresolution,
we chose $\lambda$ to be $0.001, 0.002$, and $0.006$ respectively. 

The PnP-ADMM method consists of solving the intermediate optimization problem 
\begin{equation}
    \text{argmax}_{\vx} f(\vx) + (\rho/2) \| \vx - (\vz - \vu) \|_2^2,
\end{equation}
where $\rho$ is a constant.
The values for $\rho$ we used for CT reconstruction, deblurring, and superresolution
were 0.05, 0.1, and 0.1 respectively.
We used BM3D as the denoiser with a parameter representing the noise level:
this parameter was set to 0.02 for 60 view CT and 0.05 for the other inverse problems.
A maximum of 50 iterations of conjugate gradient descent was run per outer loop.
The entire algorithm was run for 100 outer iterations at maximum
and the PSNR was observed to decrease by less than 0.005dB per iteration by the end. 

The PnP-RED method consists of the update step 
\begin{equation}
    \vx \leftarrow \vx + \mu (\nabla f - \lambda (\vx-D(\vx))),
\end{equation}
where $D(\vx)$ represents a denoiser.
The stepsize $\mu$ was set to 0.01 for the CT experiments and 1 for deblurring and superresolution.
We set $\lambda$ to 0.01 for the CT experiments and 0.2 for deblurring and superresolution.
Finally, the denoiser was kept the same as the PnP-ADMM experiments with the same denoising strength.

Table \ref{runtimes} shows the average runtimes of each of the implemented methods
when averaged across the test dataset for 60 view CT reconstruction.
\begin{table}[ht]
\centering
\begin{center}
\caption{Average runtimes of different methods across images in the test dataset for 60 view CT recon.}
\label{runtimes}
 \begin{tabular}{cc}
 \toprule
 Method & Runtime (s) $\downarrow$ \\
 \midrule
  Baseline  & 0.1 \\
 ADMM-TV  & 1 \\
 PnP-ADMM  & 73 \\
 PnP-RED  & 121  \\
 Whole diffusion  & 112  \\
 Whole SS & 248 \\
 Whole LoRA & 329  \\
 Patch diffusion & 123  \\
 Patch SS & 289  \\
 Patch LoRA & 377 \\
 \bottomrule
 \end{tabular}
 \end{center}
\end{table}

\clearpage
\subsection{Self-supervised Inverse Problem Figures}
The following figures show additional examples of self-supervised inverse problem solving. 

\begin{figure*}[ht]
\centering
\includegraphics[width=0.99\linewidth]{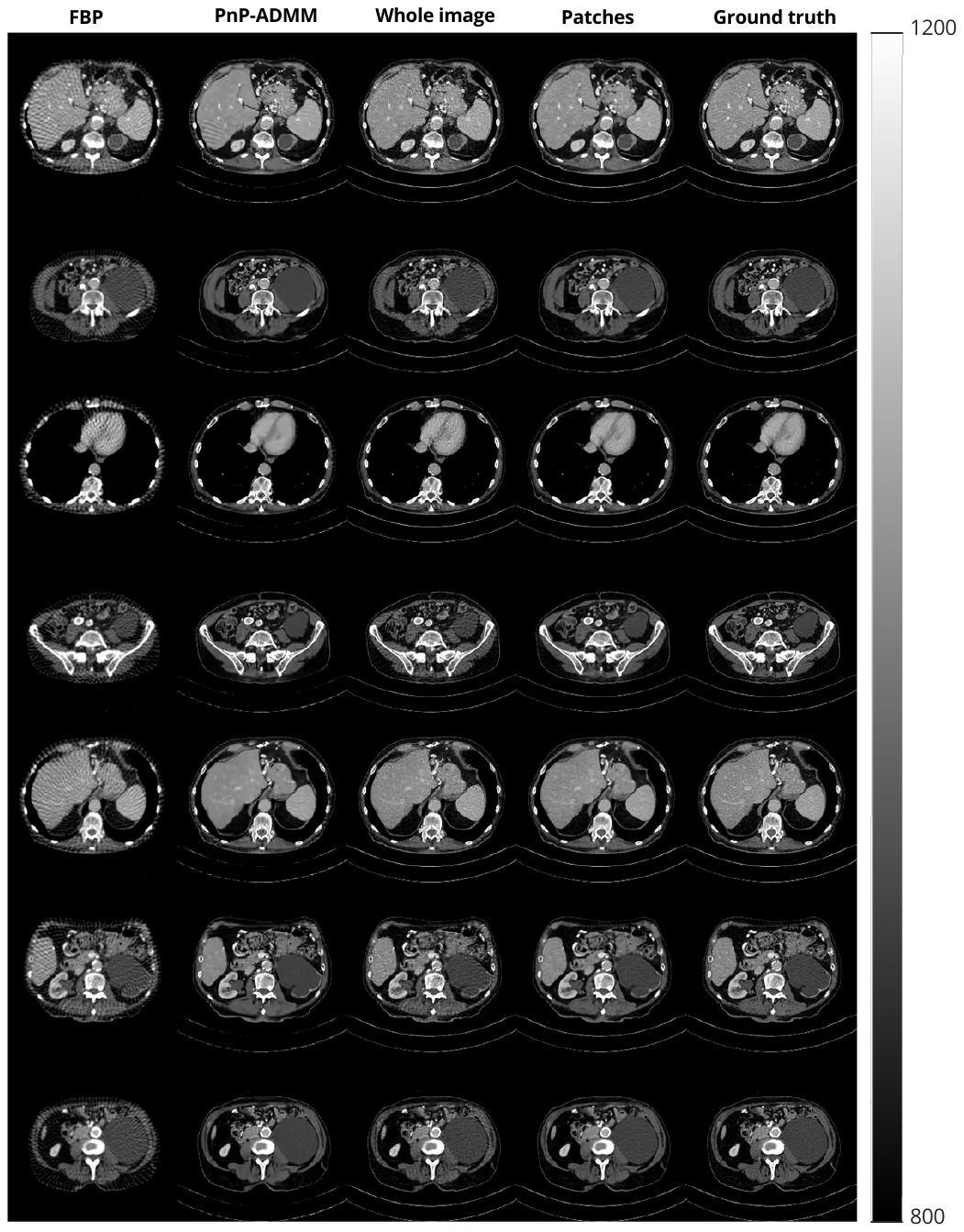}
\caption{Additional figures for self-supervised 60 view CT recon.}
\label{fig: 60view_all}
\end{figure*}

Figure \ref{fig: 60view_all} shows additional example slices of CT reconstruction from 60 views. 

\begin{figure*}[ht]
\centering
\includegraphics[width=0.99\linewidth]{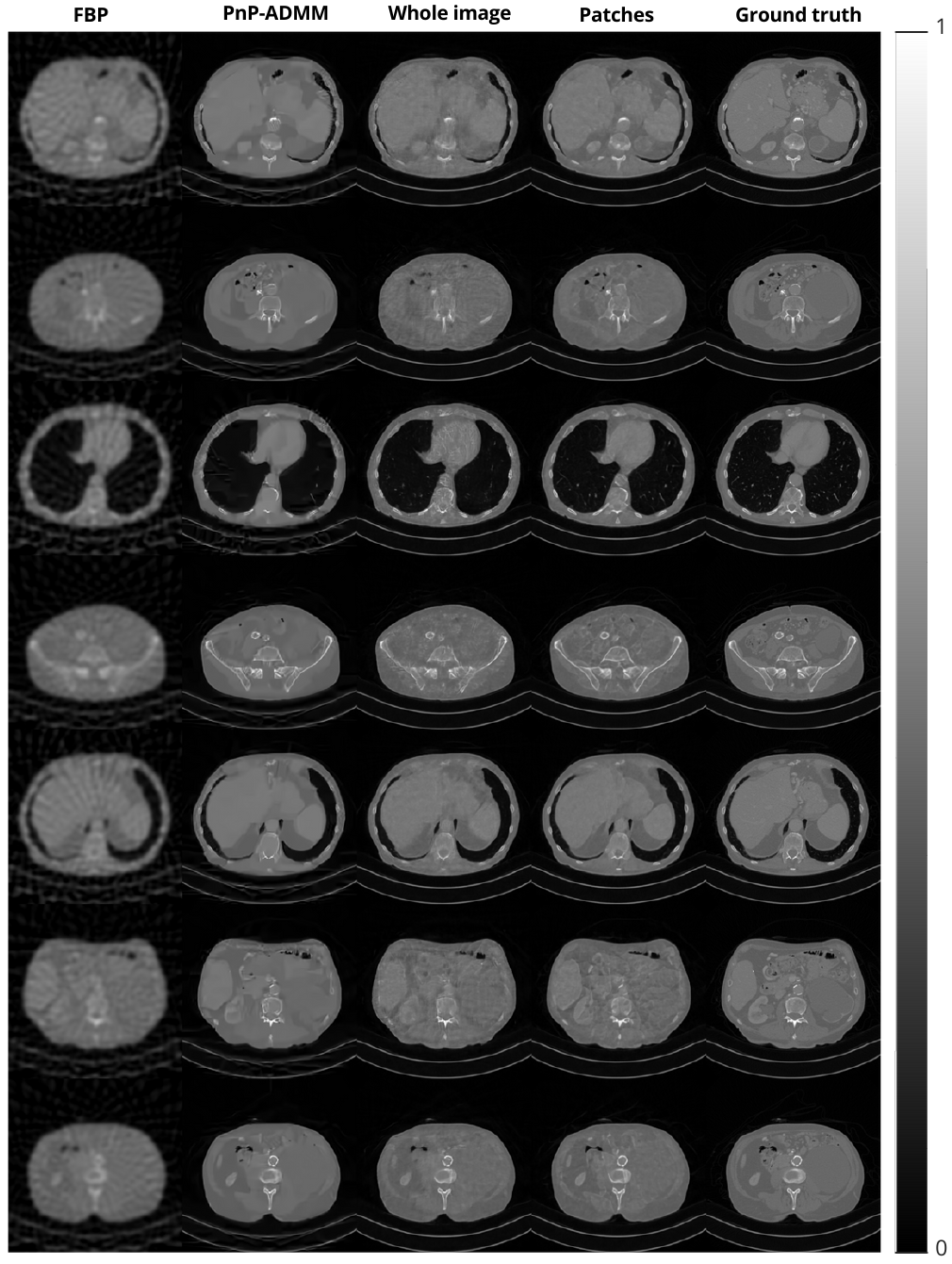}
\caption{Additional figures for self-supervised 20 view CT recon.}
\label{fig: 20view_all}
\end{figure*}

Figure \ref{fig: 20view_all} shows additional example slices of CT reconstruction from 20 views. 

\begin{figure*}[ht]
\centering
\includegraphics[width=0.99\linewidth]{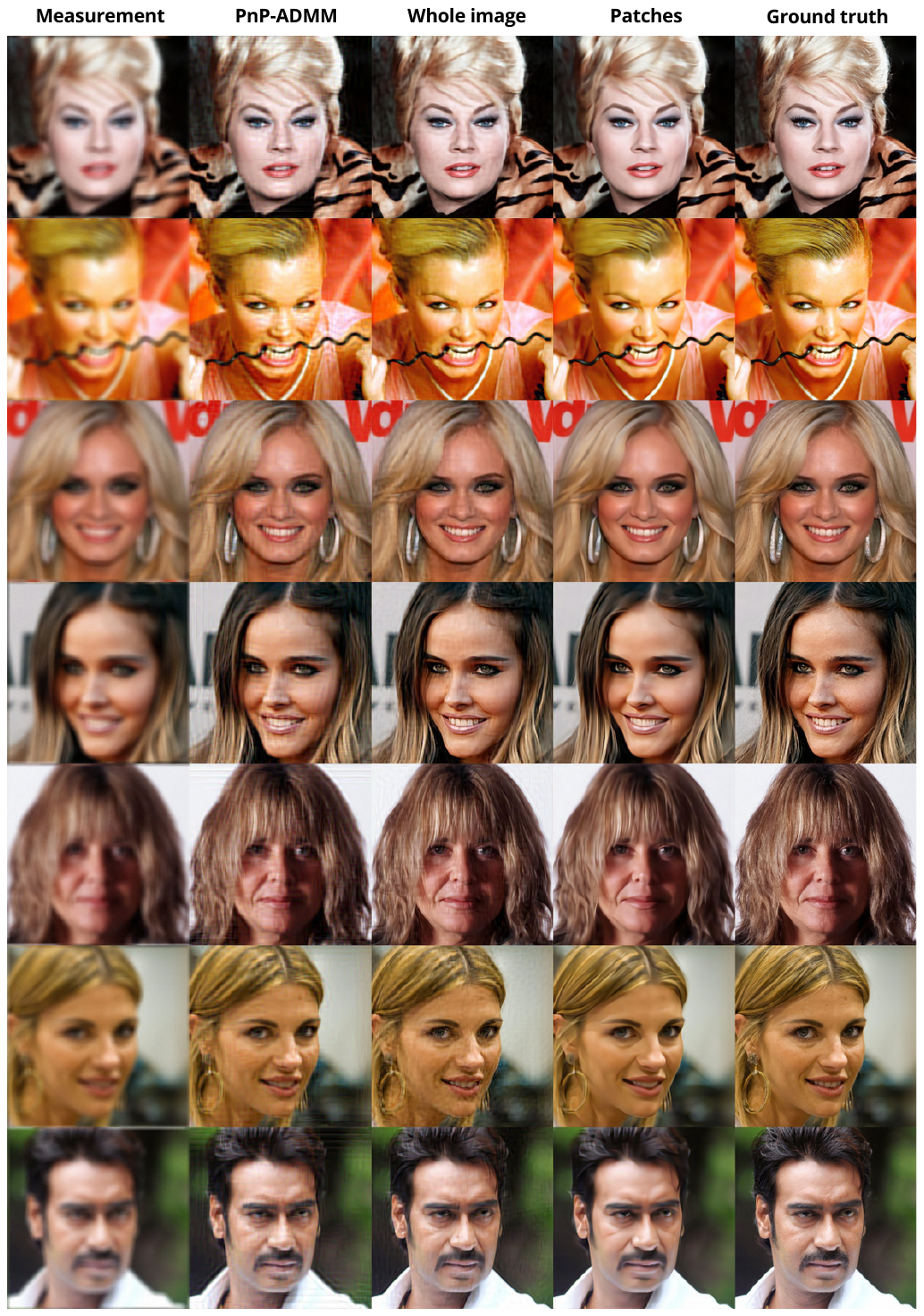}
\caption{Additional figures for self-supervised deblurring.}
\label{fig: deblur_all}
\end{figure*}

Figure \ref{fig: deblur_all} shows additional examples of deblurring with face images.

\begin{figure*}[ht]
\centering
\includegraphics[width=0.99\linewidth]{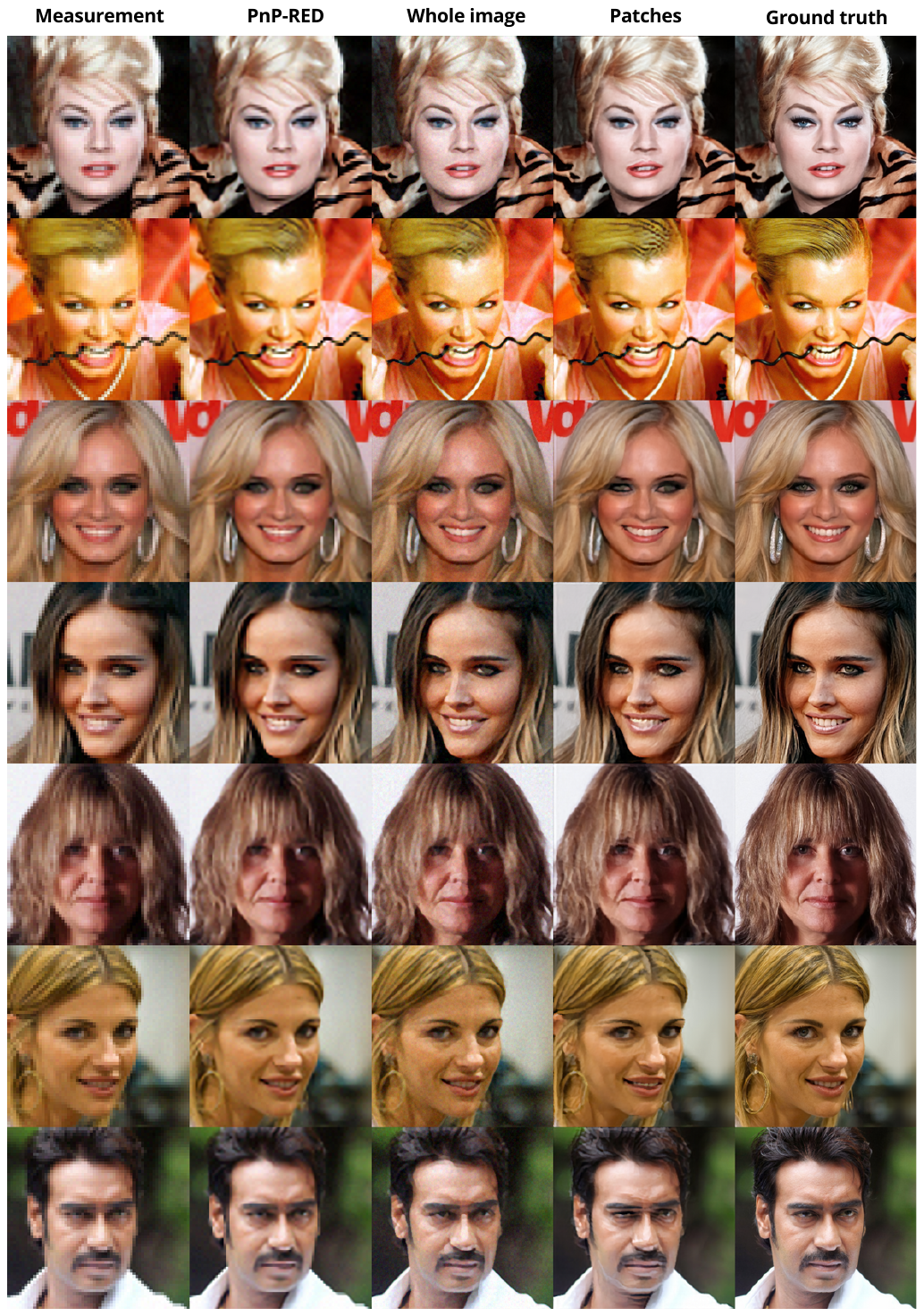}
\caption{Additional figures for self-supervised superresolution.}
\label{fig: super_all}
\end{figure*}

Figure \ref{fig: super_all} shows additional examples of superresolution with face images.

\clearpage 

\clearpage 

\clearpage 

\clearpage 

\end{document}